\documentclass{article}
\usepackage[a4paper,margin=1in,footskip=0.25in]{geometry}

\usepackage{url}
\usepackage{natbib}
\usepackage{csquotes}
\usepackage{graphicx} 
\usepackage[dvipsnames,table,xcdraw]{xcolor}
\usepackage[colorlinks=true, linkcolor=blue, citecolor=blue]{hyperref}
\usepackage{subcaption}
\usepackage{parskip}
\usepackage{booktabs} 
\usepackage{tabularx} 
\usepackage{comment}
\usepackage{graphicx}
\usepackage{subcaption}
\usepackage{amsmath}
\usepackage{wrapfig}
\usepackage{multirow}
\usepackage{makecell}
\usepackage{listings}
\usepackage{xcolor}
\usepackage[ruled,vlined]{algorithm2e}
\usepackage{mathpazo}

\def\equationautorefname~#1\null{Eq~(#1)\null}

\newif\ifdraft
\draftfalse   

\ifdraft
  \newcommand{\mansi}[1]{{\color{ForestGreen} \textbf{Mansi:} \enquote{#1}}}
  \newcommand{\colin}[1]{{\color{Blue} \textbf{Colin:} \enquote{#1}}}
  \newcommand{\marco}[1]{{\color{BurntOrange} \textbf{Marco:} \enquote{#1}}}
\else
  \newcommand{\mansi}[1]{}
  \newcommand{\colin}[1]{}
  \newcommand{\marco}[1]{}
\fi

\newcommand{\shinkaevolve}{\texttt{ShinkaEvolve}}
\newcommand{\opro}{\texttt{Modular}}
\newcommand{\openevolve}{\texttt{OpenEvolve}}
\newcommand{\optimizeanything}{\texttt{optimize\_anything}}
\newcommand{\se}{\texttt{SE}}
\newcommand{\opta}{\texttt{OA}}
\newcommand{\opene}{\texttt{OE}}

\newcommand{\promptopt}{\textbf{Prompt Optimization (GSM8K)}}
\newcommand{\po}{\textbf{Prompt Optim.}}
\newcommand{\cloud}{\textbf{CloudCast}}
\newcommand{\cant}{\textbf{Can't Be Late}}

\newcommand{\traveling}{\textbf{Traveling Salesperson ($n=100$)}}
\newcommand{\tsp}{\textbf{TSP}}

\newcommand{\circlepack}{\textbf{Circle Packing ($n=26$)}}
\newcommand{\cp}{\textbf{Circle Packing}}

\newcommand{\geminithreefive}{Gemini 3.5 Flash}
\newcommand{\geminithreeseven}{Gemini 3.7 Flash}
\newcommand{\geminitwofive}{Gemini 2.5 Flash}
\newcommand{\gptoss}{GPT-OSS (120B)}
\newcommand{\qwen}{Qwen 3.6 A3B (35B)}
\newcommand{\llama}{Llama 3.1 Instruct (8B)}

\definecolor{codebg}{rgb}{0.97,0.97,0.97}
\definecolor{codegray}{rgb}{0.5,0.5,0.5}
\definecolor{codegreen}{rgb}{0.0,0.5,0.0}
\definecolor{codepurple}{rgb}{0.58,0,0.82}

\lstdefinestyle{pseudostyle}{
    backgroundcolor=\color{codebg},
    commentstyle=\color{codegray}\itshape,
    stringstyle=\color{codegreen},
    basicstyle=\ttfamily\tiny,
    breaklines=true,
    breakatwhitespace=false,
    captionpos=b,
    keepspaces=true,
    showspaces=false,
    showstringspaces=false,
    showtabs=false,
    tabsize=4,
    frame=single,
    columns=fullflexible,
    escapeinside={(*}{*)},
}

\title{
Initialization Improves LLM-Driven Discovery

}
\author{%
    Mansi Sakarvadia\textsuperscript{\textnormal{1}}\thanks{\scriptsize Correspondence to \href{mailto:sakarvadia@uchicago.edu}{\texttt{sakarvadia@uchicago.edu}}
    },~~ 
    Marco Ciccone\textsuperscript{\textnormal{ 2}},  
    Colin Raffel\textsuperscript{\textnormal{ 2,3}}\\
    {\small \textsuperscript{1}University of Chicago,}
    {\small \textsuperscript{2}Vector Institute,}
    {\small \textsuperscript{3}University of Toronto} \\
}
\date{}

\begin{document}

\maketitle

\begin{abstract}
Large Language Models (LLMs) have been used for novel discovery of algorithms, theorems, drugs, and other tasks through the use of harnesses that prompt an LLM to iteratively optimize an objective.
In this work, we study the relationship between the population of previous iterates and eventual discovery success. 
We generalize past work on harness design to develop a suite of 12 harnesses called \opro{} and characterize their performance across 5 diverse discovery tasks,
finding that discovery success is brittle and sensitive to harness design. 
We uncover mode collapse, characterized by a dramatic drop in the diversity of iterates, as a common failure mode. 
We find that popular state-of-the-art harnesses and diversity-inducing harness interventions, which aim to prolong this collapse, yield inconsistent gains.
Our results instead uncover that the performance of early discoveries is 
predictive of eventual success.
We therefore propose a universally applicable intervention that performs an initial stage of parallel exploration in order to initialize subsequent iterative optimization.
Our method provides consistent gains across many harnesses and target applications, confirming the importance of initialization in LLM-driven discovery. 

\textbf{}
\end{abstract}

\section{Introduction} 

Large Language Models (LLMs) are increasingly playing a role in
advancing scientific and algorithmic discovery~\citep{kamatar2026empowering,woodruff2026accelerating}. 
These advances are enabled by harnessing the vast parametric knowledge of LLMs~\citep{petroni2019language, roberts2020much, geva2021transformer} by scaling \textit{inference-time compute}, i.e., spending more tokens to obtain better solutions~\citep{guo2025deepseek, muennighoff2025s1,wu2024inference}.
Recent successes include algorithm design~\citep{cheng2025barbarians}, drug repurposing~\citep{gottweis2026accelerating}, verifiable theorem proving~\citep{zheng2022miniff}, agentic system design~\citep{zhang2025aflow}, chemical hypothesis generation~\citep{summers2026chelatron}, and prompt design~\citep{agrawal2026gepa}.

While some of these discoveries have been achieved via active human-LLM teaming~\citep{woodruff2026accelerating}, there is growing interest in designing scalable, autonomous, and general-purpose \textit{discovery harnesses} \citep{novikov2025alphaevolve,openevolve,lange2026shinkaevolve,agrawal2026optimize_anything}.
These harnesses manage an LLM's in-context knowledge (i.e., prompt) as inference-time compute is spent: their primary role is to maintain a population of past discoveries which an LLM iteratively refines.
In this work, we study how populations of discoveries evolve.
In particular, we examine the factors that mediate the relationship between initial and downstream discovery quality.

\begin{figure}[b!t]
    \centering
    \includegraphics[width=\textwidth]{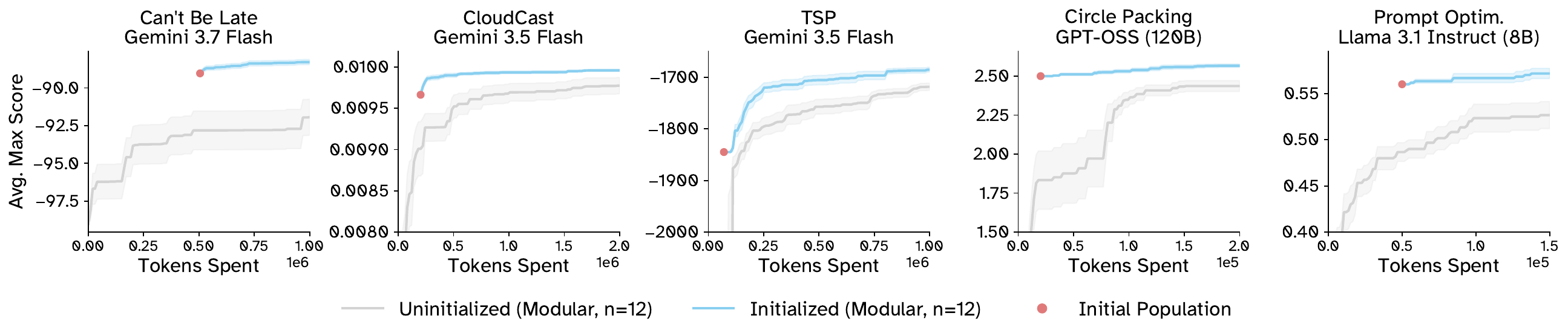}
    \caption{\textbf{Initialization raises the floor for discovery under a fixed total token budget.} We find that initializing (the \opro{}) discovery harnesses with a high-performing initial population is beneficial for downstream discovery success across harnesses, tasks, and LLMs.
    Initialization budget is informed by the convergence criteria in~\autoref{sec:initialization}. 
    Initial population size $m=15$, similarity threshold $s=0.98$. 
    }
    \label{fig:initialized_trajectories}
\end{figure}


We begin by validating the role of in-context knowledge as a key player in discovery quality. We do so by expanding prior work on harness design to develop a suite of 12 discovery harnesses, referred to as \opro{}, that can expose a wide range of harness behavior by varying design choices. 
We use these harnesses to study whether sequentially refining an initial seed discovery is beneficial.
To answer this, we directly compare the discovery performance of sequential harnesses in \opro{}, which continuously update the LLM's context with knowledge from recent discoveries, against a simpler approach: repeatedly querying the LLM in parallel to refine the initial seed without updating the prompt with new iterates. We control the inference-time compute expended by each discovery strategy and find that parallel prompting beats \textit{some} but not \textit{all} of the discovery harnesses. Across five diverse discovery tasks, the top-performing harnesses differed, leading us to conclude that harnessed LLM-guided discovery, while promising, is \textit{brittle}. This means that in-context knowledge is crucial to downstream discovery success, but how that in-context knowledge is best curated and managed is both task- and LLM-specific. 


Despite the variability in harness performance across different task/LLM settings, we find, on average, that harnesses experience 
diminished discovery diversity over the duration of the trajectory, which we refer to as \textit{mode collapse}.
Therefore, we examine if harnesses that \textit{explore} more (i.e., generate more diverse discoveries) are
beneficial for downstream discovery.
Specifically, we consider two approaches to enable more diverse discovery: 1) state-of-the-art harnesses that prioritize diverse discoveries 
and 2) diversity-oriented interventions to manage discovery harness populations.
Once again, we find that ideal harness design and interventions are sensitive to the task and choice of LLM. Therefore, no approach consistently enables trajectories to recover from poor initial discoveries. However, a commonality of the highest performing harnesses across tasks and LLMs is that they enable a performant initial discovery population.
Spurred by this insight, \textbf{we confirm that the performance of early iterates can predict of downstream discovery success.}

Therefore, we hypothesize that explicitly initializing a discovery harness with a high-performing population will benefit discovery.
To test this, we propose a simple task-, LLM-, and harness-invariant initialization method: use a portion of the total compute budget to generate a pool of initial iterates through a parallel optimization stage.
We then warm-start sequential discovery harnesses with the high-performing candidates from this initial population.
In so doing, we observe that, even after controlling for initialization compute, our approach can beat both uninitialized sequential and parallel discovery. 
Additionally, we establish a simple method to determine the initialization budget: use a simple convergence criterion to decide when to curate a high-performing initial population from the parallel pool and begin sequential optimization. 
This convergence criterion consistently predicts a near-ideal initialization budget and subsequently outperforms purely sequential and parallel compute across task-, LLM-, discovery harnesses.
Finally, we validate that explicitly initializing state-of-the-art harnesses with our method consistently increases discovery performance by 2-21\%.

\section{Background on LLMs as Optimizers}
\label{sec:background}
Here we introduce the practice of using LLMs as optimizers for discovery tasks, enumerate the general design principles that govern discovery harnesses, describe the metrics we use to quantitatively compare different discovery trajectories, and outline the discovery tasks we consider in this study.

\subsection{Sequential Discovery Harnesses}
\label{sec:discovery_harness}

\cite{yang2024large} proposed to use LLMs to sequentially optimize an objective expressed in natural language by iteratively prompting an LLM with the objective and examples of past solutions. \textit{Discovery harnesses} instantiate this general workflow to enable better downstream discovery (\autoref{appendix:discovery_harness}, \autoref{alg:discovery-harness}). Generally, a harness \textit{samples} parent discoveries from an \textit{active population} of past discoveries. An LLM is then prompted to \textit{mutate} these parents to generate an improved \textit{discovery}, with respect to the objective. New discoveries undergo task-specific \textit{evaluation} and are added to the active population if they meet certain criteria. The active population is periodically \textit{pruned} when it reaches capacity. 
These harnesses can be crafted in a task-specific manner~\citep{guo2024connecting,agrawal2026gepa,zhang2025aflow}, but interest in \textit{task-agnostic} discovery harnesses is growing~\citep{openevolve,agrawal2026optimize_anything,maheswaran2026squeeze,lange2026shinkaevolve}. Here, we study the drivers of failures and successes of these general-purpose discovery harnesses, and therefore consider a wide range of harnesses with differing design choices.

\textbf{\opro{} Harnesses:} We develop a set of harnesses called \opro{} by generalizing the \texttt{OPRO} discovery framework proposed by~\cite{yang2024large}. We minimally modify the \texttt{OPRO} harness to expose two axes of freedom: 1) the LLM mutation strategy and 2) the parent sampling strategy to enable a broader range of harness 
behavior. We consider four mutation strategies (\autoref{lst:kincontext-mutator}-\ref{lst:ga-mutator}) and three sampling strategies (\autoref{alg:high-score-sampling}-\ref{alg:roulette-wheel-sampling}), resulting in a suite of 12 unique harnesses detailed in Appendix~\ref{appendix:modular_harnesses}.

\textbf{State-of-the-art (SOTA) Harnesses:} We consider three popular, well-engineered discovery harnesses from the recent literature: \shinkaevolve{} (\se{})~\citep{lange2026shinkaevolve}, \openevolve{} (\opene{})~\citep{openevolve}, and \optimizeanything{} (\opta{})~\citep{agrawal2026optimize_anything} as they are widely adopted and studied. We detail their hyper-parameter settings in Appendix~\ref{appendix:prod_grade_harnesses}. 

\subsection{Discovery Metrics}
\label{sec:background_metric}
We outline metrics to quantify a discovery scheme that generates $D=\{d_i,...,d_n\}$ discoveries.

\textbf{Maximum Score:} In a discovery trajectory, all discoveries $D$ are scored using a task-specific evaluation function (detailed in~\autoref{tab:discovery_task_descriptions}). We follow the standard practice to define the quality of a discovery trajectory by the maximal score achieved by any discovery within that trajectory. 

\textbf{Tokens Expenditure:} We quantify the cost of a discovery trajectory by the number of tokens used to generate all discoveries $D$. Past work has quantified the cost of a trajectory by the number of discovery evaluations~\citep{agrawal2026gepa,agrawal2026optimize_anything,lange2026shinkaevolve,openevolve}; however, this is not an exact comparison as some discovery schemes expend more inference-time compute to generate and evaluate a single discovery. We therefore measure token expenditure instead. 

\textbf{Diversity:} Following~\citet{wenger2026large}, we measure the semantic similarity between two LLM-generated discoveries $d_1, d_2$ by embedding them with a sentence embedding model and computing the cosine similarity between their embeddings. We extend this approach to quantify the \textit{diversity} of a set of discoveries $D$, following~\citet{zhang2024improving}. Letting $E = \{e_1, \dots, e_n\}$ denote $D$'s corresponding embeddings, $\text{Diversity}(D) = 1 - \overline{\text{sim}}(E)$, where $\overline{\text{sim}}(E)$ is the average pairwise cosine similarity over all $e_i, e_j \in E$, $i \neq j$. We use a general embedding model (\texttt{all-MiniLM-L6-v2}) for natural language tasks (as per~\cite{wenger2026large}) and a code embedding model (\texttt{nomic-ai/CodeRankEmbed}) for coding tasks.

\subsection{Discovery tasks}

We consider five optimization tasks that cover a wide range of domains from~\citep{cheng2025barbarians,agrawal2026optimize_anything,yang2024large}. We briefly describe them below and further detail them, their LLMs, token budgets, and evaluation metrics in~\autoref{appendix:discovery_task_description},~\autoref{tab:discovery_task_descriptions}:

\begin{enumerate}
    \item \cloud{}: Discover an algorithm to optimize multi-cloud data transfer cost~\citep{wooders2024cloudcast}.
    \item \cant{}: Discover an algorithm for single spot instance deadline-driven job scheduling~\citep{wu2024can}.
    \item \circlepack{}: Discover an algorithm to arrange 26 circles of variable diameters inside a unit square without overlap such that cumulative diameters are maximized.
    \item \traveling{} (\tsp{}): Discover the shortest path that visits 100 cities (with predefined inter-city distances) exactly once and returns to the starting point.
    \item \promptopt{}: Discover an instruction prompt that enables a downstream LLM to maximize score on the GSM8K benchmark~\citep{cobbe2021training}.
\end{enumerate}

We repeat experiments for all tasks over multiple LLMs and report a subset of these experiments in the main text (see \autoref{tab:main_task_llm_pairing}); full experimental results are reported in~\autoref{appendix:extended_results}.


\section{
Parallel vs. Sequential Discovery
}
\label{sec:parallel_v_seq}


\begin{figure}[b!t]
    \centering
    \includegraphics[width=\textwidth]{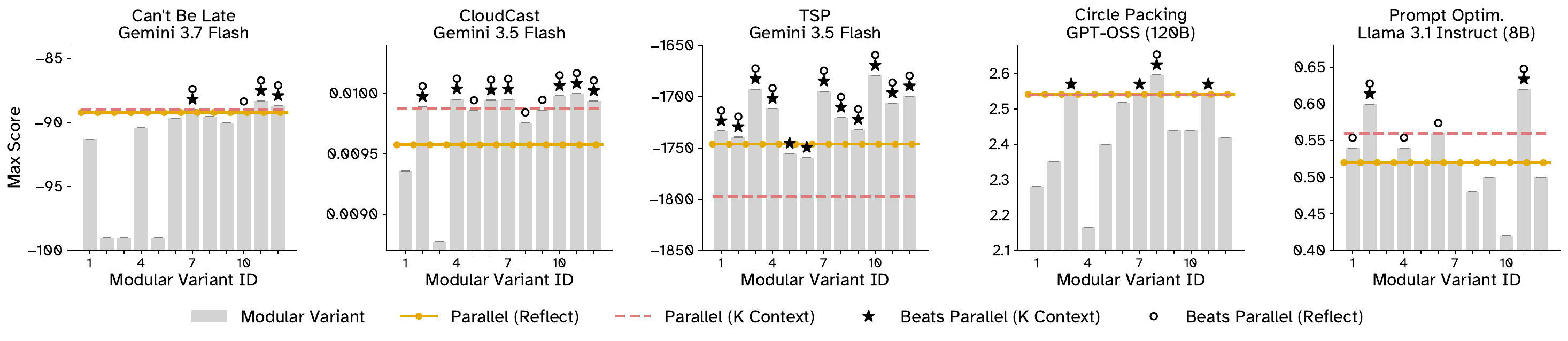}
    \caption{\textbf{Sequential discovery can outperform parallel discovery.} Comparing sequential  discovery harnesses from our \opro{} suite (detailed in~\autoref{tab:opro_variants}) with parallel discovery using the ``Reflect'' (\autoref{lst:reflect-mutator}) and ``K Context'' prompts (\autoref{lst:kincontext-mutator}). Higher scores are better. Both parallel and sequential approaches are initialized with the same initial seed solutions. We see that generally some form of sequential compute outperforms parallel; but the types of harnesses that perform the best vary with task and LLM. Full results in~\autoref{fig:seq_v_parallel_full_grid}.}
    \label{fig:parallel_v_sequential}
\end{figure}

Here we ask the question: \textit{How should inference-time compute be allocated and managed to enable better discovery?} Sequential \textit{discovery harnesses}, which serially query a language model to refine a discovery, conditioned on past iterates, have become popular~\citep{yang2024large, novikov2025alphaevolve, openevolve, lange2026shinkaevolve,agrawal2026optimize_anything}. We seek to validate the implicit assumption in these works that sequential refinement of a discovery is better than a simple alternative: directly querying the LLM independently in parallel~\citep{brown2024large, wang2026rethinking}. 

\begin{wraptable}{r}{0.5\textwidth}
\vspace{-0.5cm}
\centering
\begin{tabular}{lc}
\toprule
    Task & Optimizer LLM \\ 
    \midrule
    \cloud{} & \geminithreefive{}\\
    \cant{} & \geminithreeseven{} \\
    \tsp{} & \geminithreefive{} \\
    \cp{} & \gptoss{} \\
    \po{} & \llama{}\\
\bottomrule
\end{tabular}
\caption{A subset of (task, LLM) pairs studied. Full set detailed in~\autoref{appendix:discovery_task_description},~\autoref{tab:discovery_task_descriptions}.}
\label{tab:main_task_llm_pairing}

\end{wraptable}

We begin with a straightforward comparison: for each (discovery task, LLM) in~\autoref{tab:main_task_llm_pairing}, we compare the highest-scoring discovery across the 12 sequential \opro{} harnesses (see~\autoref{tab:opro_variants}) against parallel discovery.  In both sequential and parallel settings, the initial seed iterate and token budget are the same (\autoref{appendix:discovery_task_description}). In parallel discovery, the initial seed iterate is mutated in parallel by an LLM prompted with the task's discovery objective until the token budget is exhausted. Here, the LLM's context never changes; we study this parallel approach using two different mutator prompts (i.e., Reflect and K Context; see Appendix~\ref{appendix:parallel_discovery}). Sequential harnesses similarly employ an LLM to mutate discoveries, but instead iteratively update the LLM's context with information from recent discoveries (see Appendix~\ref{sec:discovery_harness}).

In~\autoref{fig:parallel_v_sequential}, we plot the maximal score achieved by all \opro{} harness variants and parallel discovery schemes. 
Since parallel discovery never updates the LLM's context, its performance is dependent on the LLM's parametric knowledge base. In parallel discovery, the ceiling for discovery performance is not known a priori, as reliably measuring the LLM's parametric knowledge for discovery tasks is challenging. For example, in~\autoref{fig:parallel_v_sequential}, we observe that for \tsp{}, parallel discovery typically underperforms sequential \opro{} harnesses, while for \cp{} the trend is reversed. On the other hand, sequential harnesses enable LLMs to build on the in-context knowledge from past discoveries, rather than relying solely on parametric knowledge.
For each task, we observed that multiple \opro{} harnesses outperformed parallel discovery, 
but that the specific outperforming \opro{} variants vary. 
This suggests that sequential refinement of a discovery can outperform parallel optimization but the best sequential harness design changes per (task, LLM) pairing, indicating that sequential discovery is sensitive to harness hyper-parameter choices. We replicate these findings for each task over additional LLMs in~\autoref{appendix:extended_results},~\autoref{fig:seq_v_parallel_full_grid}.

\begin{figure}[htbp]
    \centering
    \includegraphics[width=\textwidth]{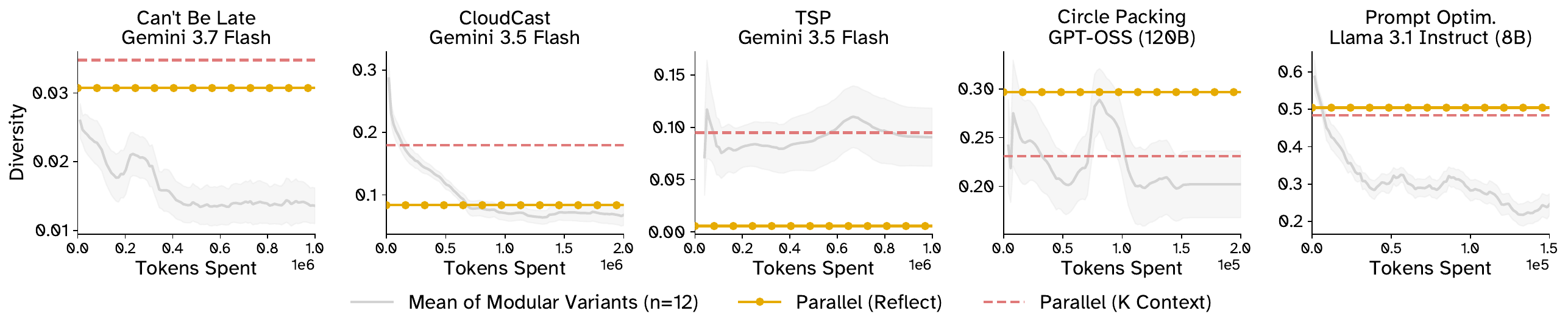} 
    \caption{\textbf{Sequential discovery is prone to mode collapse.}  The windowed average pairwise diversity of discoveries over the course of a trajectory across all \opro{} variants vs. average diversity of parallel discovery. Diversity for the parallel discovery scheme is computed over the full set of discoveries. Sequential harnesses consistently yield fewer diverse discoveries and exhibit diminished diversity over the trajectory, whereas parallel schemes discover more diverse candidates. Window size = 15 for all tasks except \cp{} which has window size = 6. Full results in~\autoref{fig:mode_collapse_grid}.}
    \label{fig:mode_collapse}
\end{figure}

To better understand failure modes of sequential discovery, we analyze the populations of iterates between sequential and parallel discovery schemes. In~\autoref{fig:mode_collapse}, we plot the diversity of the entire population of discoveries from a parallel trajectory vs. the windowed diversity over a sequential trajectory averaged across all \opro{} harness variants. We find that the parallel schemes generally have more diverse populations of discoveries than sequential. Further, we characterize \textit{mode collapse} in sequential trajectories: the diversity of new candidates decreases over the course of a sequential trajectory. We hypothesize that this is due to the in-context prior, which is present and reinforced throughout sequential discovery, but absent in parallel discovery. Despite mode collapse,~\autoref{fig:parallel_v_sequential} demonstrates that sequentially refining a seed discovery can be beneficial and can outperform parallel discovery if the sequential harness discovers a high-performing mode.



\section{
Can Exploration Enable Better Discovery?
}
\label{sec:online_intervention}

Predicated on the finding from~\autoref{sec:parallel_v_seq} that sequential discovery harnesses are prone to mode collapse, we ask: \textit{Can additional exploration enable discovery trajectories to recover from low-performing modes?}
We use the diversity of the discovered population as a proxy metric for how much a discovery scheme \enquote{explored}.
We consider two different approaches to assess the effects of exploration on discovery quality:
\begin{enumerate}
    \item We compare our simple \opro{} harnesses to a set of state-of-the-art harnesses that treat population diversity as a key consideration in harness design. 
    \item We control for harness design variation by applying a class of online interventions to \opro{} harnesses that aim to maintain diversity as per~\citep{lange2026shinkaevolve}. 
\end{enumerate}

\begin{table}[b]
\centering
\scriptsize
\begin{tabular}{llllll}
\toprule
 & \cant{} & \cloud{} & \tsp{} & \cp{} & \po{} \\
\midrule
\opro{} (n=12) & 0.02 ± 0.01 & 0.1271 ± 0.0195 & 0.095 ± 0.050 & 0.26 ± 0.06 & 0.40 ± 0.04 \\
\midrule
\opro{} w/ Diversity Interven. (n=60) & 0.06 ± 0.02 & 0.1417 ± 0.0132 & 0.100 ± 0.022 & 0.27 ± 0.02 & 0.38 ± 0.03 \\
State-of-the-art (n=3) & 0.05 ± 0.07 & 0.1680 ± 0.3698 & 0.259 ± 0.332 & 0.25 ± 0.10 & 0.36 ± 0.23 \\
\bottomrule
\end{tabular}
\caption{\textbf{Discovery harnesses can discover more diverse populations.} We compute the diversity of all discoveries in a trajectory, and then report the average diversity across different harnesses. Both state-of-the-art discovery harnesses (\se{}, \opta{}, \opene{}) and \opro{} harnesses with diversity interventions on average generally achieve more diverse discoveries than baseline~\opro{} harnesses. Full results in~\autoref{appendix:extended_results},~\autoref{tab:diversity_of_harnesses_all}.}
\label{tab:diversity_of_harnesses_summary}
\end{table}

\begin{table}[h]
\centering
\scriptsize
\begin{tabular}{llllll}
\toprule
 & \cant{} & \cloud{} & \tsp{} & \cp{} & \po{} \\
\midrule
\opro{} (n=12) & -91.94 ± 2.75 & \textbf{0.0098 ± 0.0002} & \textbf{-1719 ± 16} & \textbf{2.44 ± 0.08} & \textbf{0.53 ± 0.03} \\
\midrule
Sim thresh. (0.95) (n=12) & -95.79 ± 2.98 & 0.0097 ± 0.0002 & -1777 ± 29 & 2.19 ± 0.38 & \textbf{0.53 ± 0.03} \\
Sim thresh. (0.8) (n=12) & \textbf{-91.68 ± 2.81} & 0.0097 ± 0.0001 & -1775 ± 17 & 2.34 ± 0.26 & 0.52 ± 0.02 \\
\midrule
Sim thresh. (0.95) + LLM-Judge (n=12) & -95.93 ± 2.82 & \textbf{0.0098 ± 0.0001} & -1744 ± 46 & 2.33 ± 0.23  & \textbf{0.53 ± 0.04} \\
Sim thresh. (0.8) + LLM-Judge (n=12) & -94.82 ± 2.75 &  \textbf{0.0098 ± 0.0001} & -1735 ± 37 & 2.26 ± 0.30 & 0.51 ± 0.04 \\
\midrule
Deduplication (n=12) & -93.93 ± 3.34 & \textbf{0.0098 ± 0.0001} & -1721 ± 12 & 2.04 ± 0.44 & 0.52 ± 0.04 \\
\bottomrule
\end{tabular}
\caption{\textbf{Diversity interventions do not consistently improve discovery harness performance.} Each cell contains the average maximal score achieved over a set of harnesses. The basic~\opro{} harnesses generally have the highest discovery quality, with some diversity interventions matching, and rarely exceeding, performance. 
Sim thresh. is the cosine similarity threshold ($s$). Bold indicates highest score within a column. Full results in~\autoref{appendix:extended_results},~\autoref{tab:online_interventions_compare_all}.
}
\label{tab:online_diversity}
\end{table}

\textbf{State-of-the-art Harnesses.} 
We directly compare the performance of the state-of-the-art \shinkaevolve{} (\se{}), \openevolve{} (\opene{}), and \optimizeanything{} (\opta{}) harnesses to \opro{} variants across the same (task, LLM) pairs as in ~\autoref{sec:parallel_v_seq}; all harnesses share the same initial seed iterate. In~\autoref{tab:prod_vs_modular_harnesses}, we compare the average maximum discovery score across \opro{} and state-of-the-art harnesses. Despite validating that the more sophisticated harnesses do, on average, generate more diverse discoveries (i.e., explored more, see~\autoref{tab:diversity_of_harnesses_summary}), we find that they often underperform less diverse \opro{} harnesses. This suggests that harness design choices, which interact differently with the task and LLM, rather than exploration, are a primary mediator of successful discovery. 

\begin{table}
\centering
\scriptsize
\begin{tabular}{llllll}
\toprule
 & \cant{} & \cloud{} & \tsp{} & \cp{} & \po{} \\
\midrule
\opro{} (n=12) & -91.94 ± 2.75 & \textbf{0.0098 ± 0.0002} & \textbf{-1719 ± 16} & 2.44 ± 0.08 & \textbf{0.53 ± 0.03} \\
State-of-the-art (n=3) & \textbf{-91.32 ± 8.33} & 0.0095 ± 0.0021 & -1772 ± 250 & \textbf{2.51 ± 0.07} & 0.47 ± 0.20 \\
\bottomrule
\end{tabular}
\caption{\textbf{State-of-the-art discovery harnesses (\se{}, \opta{}, \opene{}) do not consistently outperform \opro{} harnesses.} The ideal set of discovery harnesses varies across (task, LLM) settings. Bold indicates highest score within a column. Full results in~\autoref{appendix:extended_results},~\autoref{tab:online_interventions_compare_all}.}
\label{tab:prod_vs_modular_harnesses}
\end{table}

\begin{figure}[tbp]
    \centering
    \includegraphics[width=\textwidth]{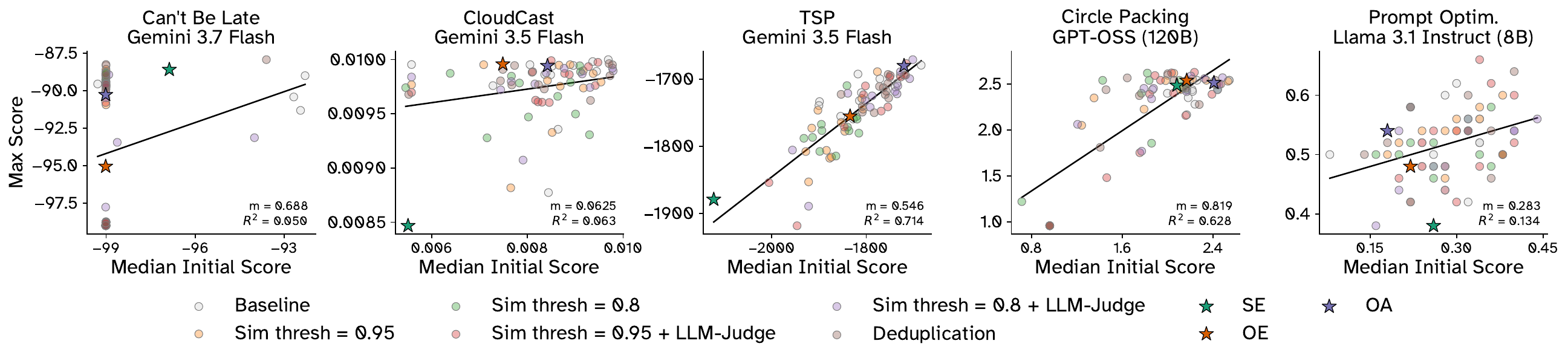}
    \caption{\textbf{Initial discovery quality can predict downstream quality.}  We plot the correlation between the median score of the first 15 discoveries vs. the max score achieved in a trajectory across varying discovery harnesses (72~\opro{} harnesses and 3 state-of-the-art harnesses).}
    \label{fig:correlations}
\end{figure}

\textbf{Online Diversity Interventions.}
Next, to more precisely measure the effect of exploration on discovery quality, we directly apply an intervention on the \opro{} harnesses while controlling for the effect of harness design.
Following~\cite{lange2026shinkaevolve}, we consider a class of online methods that intervene on the active population and aim to maintain a diverse population: rejecting a newly discovered $d_{\text{new}}$ from entering the active population $P = \{d_1,...,d_m\}$  if $d_{\text{new}}$ is too similar to current candidates, where $m$ is the max population size. We consider three simple and popular online diversity interventions:
\begin{enumerate}
    \item \textbf{Cosine Similarity Threshold}~\citep{lange2026shinkaevolve}: 
    Reject $d_{\text{new}}$ if its embedding exceeds the cosine similarity threshold $s$ to the embedding of any candidate already in $P$. We use the sentence embedding models detailed in~\autoref{sec:background_metric} to embed candidates.
    \item \textbf{LLM-Judge}~\citep{lange2026shinkaevolve}: Identify the most similar candidate $d^*$ to $d_{\text{new}}$ from $P$ by comparing cosine similarities of the candidates' embeddings. Reject $d_{\text{new}}$ if its cosine similarity to $d^*$ exceeds $s$ \textit{and} an LLM-Judge (\geminithreefive{}) does not deem $d_{\text{new}}$ meaningfully different from $d^*$. See Appendix~\ref{appendix:llm_judge} for the LLM-Judge prompt.
    

    \item \textbf{Deduplication:} Reject $d_{\text{new}}$ if $d_{\text{new}} \in P$.
\end{enumerate}

\begin{figure}[t]
    \centering
    \includegraphics[width=\textwidth]{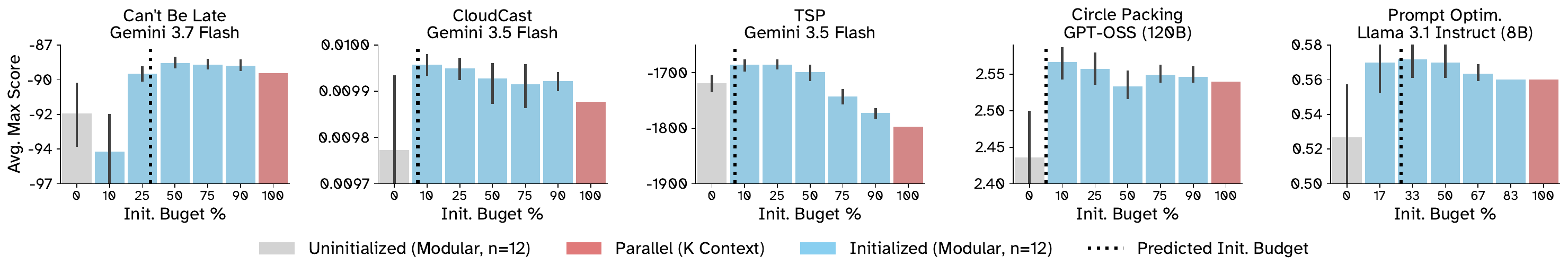}
    \caption{\textbf{Initialized discovery can outperform purely sequential and parallel discovery.} Sequential discovery harnesses initialized with a high-performing population (blue) via a fraction of the discovery budget yield the most performant discoveries.
    The ideal initialization to total discovery budget ratio (dotted line) can be successfully predicted via a convergence analysis~\autoref{sec:initialization}. Diversity threshold $s=0.98$ for all initial populations, population size $m=15$. Full results in~\autoref{fig:initialization_compute_grid}.
    }
    \label{fig:initialization_compute}
\end{figure}

We run the same set of sequential experiments with the \opro{} harness variants as~\autoref{sec:parallel_v_seq}, but we vary the diversity interventions: cosine similarity thresholding with $s\in\{0.95, 0.8\}$, LLM-Judge with $s\in\{0.95, 0.8\}$, and deduplication. We validate that these interventions do generally enable harnesses to explore more and achieve more diverse discoveries (\autoref{tab:diversity_of_harnesses_summary}). However, these interventions have variable effects on discovery quality and, in some cases, harm performance (\autoref{tab:online_diversity}). 
In some cases, exploration negatively impacts convergence towards high-performance discoveries. Further, we find that the interventions that enable better discovery vary across tasks and are sensitive to hyper-parameters.

\begin{wrapfigure}{r}{0.35\textwidth}
    \centering
    \vspace{-\intextsep}
    \includegraphics[width=0.35\textwidth]{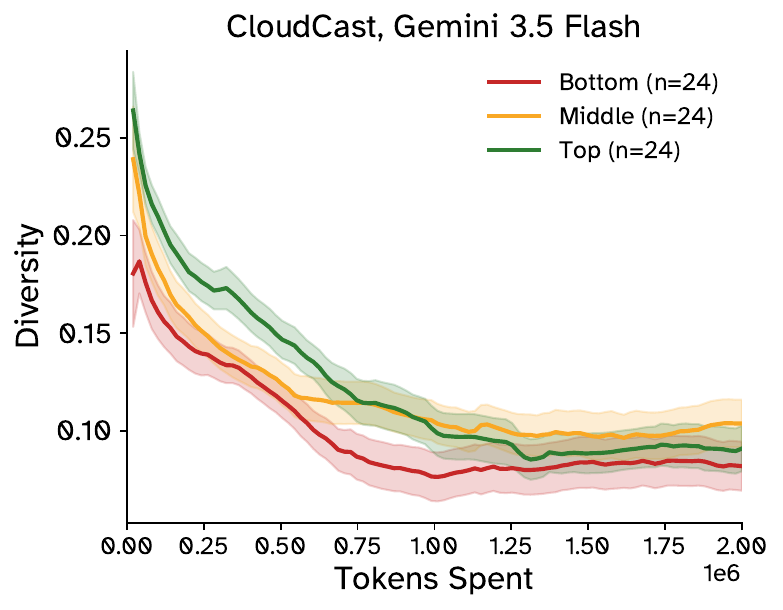}
    \caption{\textbf{High scoring trajectories have higher initial diversity but eventually experience mode collapse.} We stratify the \opro{} harness trajectories (for \cloud{}) from~\autoref{tab:online_diversity} into tertiles based on the max score of a given trajectory. Windowed diversity visualized over trajectory (window size=15). Full results in~\autoref{appendix:extended_results},~\autoref{fig:modecollapse_grid}. }
    \label{fig:cloudcast_modecollapse}
    \vspace{-\intextsep}
\end{wrapfigure}

\textbf{High-performing discovery trajectories can collapse onto favorable modes.} Our results from state-of-the-art harnesses and diversity interventions show that when harnesses start from the same seed, those that explore more (i.e., maintain more diverse populations) do not reliably produce better-performing discoveries. 
For example, in~\autoref{fig:cloudcast_modecollapse}, we visually stratify all 72 discovery trajectories for the~\cloud{} task (from~\autoref{tab:online_diversity}) into tertiles by their best score. We observe that while high-performing trajectories can be marginally more diverse early on, they too can experience mode collapse over time. 
This suggests a more nuanced conclusion: in a trajectory with poor initial candidates, mode collapse around those candidates can prevent better ones from ever being discovered, so intervening on the population can help. In contrast, an initially high-performing trajectory may actually be hurt by unnecessary exploration and benefit from more exploitation. Distinguishing between these two scenarios is challenging, and \textbf{exploration does not guarantee recovering from poor initial discoveries.}

In~\autoref{fig:correlations}, we visualize the relationship between the initial score of the first 15 discoveries in a discovery trajectory vs.\ the maximum downstream score achieved by that trajectory. While no specific harness consistently performed well, the highest-performing harnesses for each (task, LLM) pairing tended to be those that produced a high-performing initial population. \textbf{We find that the quality of the initially discovered population can be predictive of downstream discovery quality in a trajectory. }


\vspace{1em}

\section{The Importance of Initialization
}
\label{sec:initialization}

Since the eventual success of a sequential trajectory can be correlated with the performance of the first few discovered candidates (\autoref{fig:correlations}), we posit that 
it is beneficial to initialize a high-performing population of candidates that can better condition downstream sequential optimization. 
We ask the question: \textit{can we better allocate compute so that the performance of the initial candidate population is boosted?} 

Given a discovery budget, we propose a method to stratify compute into a population initialization phase followed by a serial discovery phase. 
We find that by spending a fraction of the compute budget ($r$) on parallel discovery, we can curate initial populations that are more amenable to downstream refinement via sequential optimization.
This produces a simple and extensible initialization method that can be adopted with \textit{any} sequential discovery harness: first generate a pool of discoveries using parallel discovery; then curate a performant population $p_{\text{init}}$ of size $m$ by greedily selecting candidates from that parallel pool.
We apply cosine similarity thresholding when greedily initializing a population (see description in~\autoref{sec:online_intervention} \&~\citep{lange2026shinkaevolve}).
If we exhaust sufficiently unique candidates before the population has reached size $m$, we greedily populate the remaining population slots with the deduplicated highest performers from the parallel pool. 

We repeat the set of experiments on \opro{} harnesses from ~\autoref{sec:parallel_v_seq} but with five different initialization budgets ($r\in[0,1]$) per task (see task-specific ratios in~\autoref{appendix:discovery_task_description},~\autoref{tab:init_ratios}). We let the initial population size $m=15$ and consider similarity thresholds $s\in\{1, 0.98, 0.95, 0.8\}$ where $s=1$ corresponds to greedy sampling. In~\autoref{fig:initialization_compute}, we plot the average maximal score achieved across all 12 \opro{} variants across different initialization budgets; $r=0$ means sequential trajectories with no initialization and $r=1$ means fully parallel discovery. We observe that our parallel initialization scheme uniformly outperforms both fully serial and parallel optimization.

\begin{figure}[t]
    \centering
    \includegraphics[width=\textwidth]{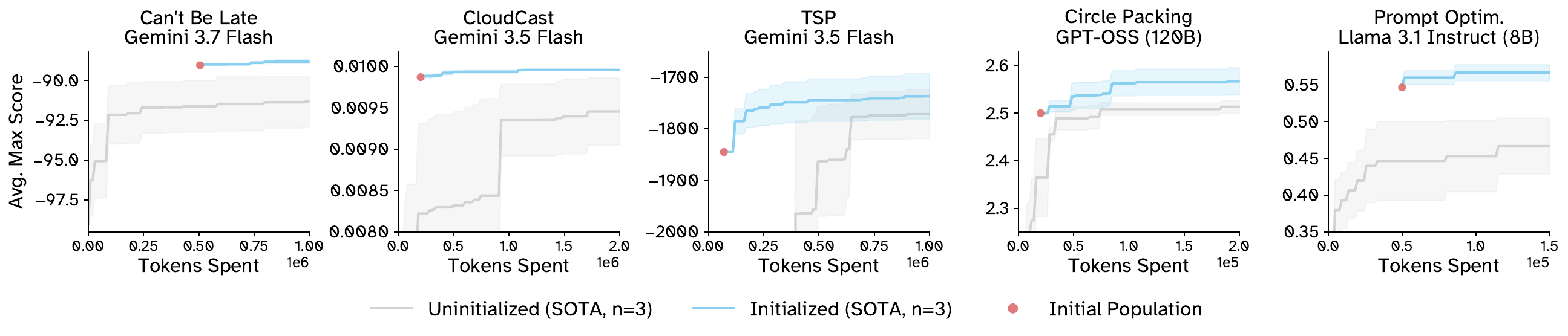}
    \caption{\textbf{Initializing state-of-the-art harnesses improves discoveries.} 
    We find that initializing the state-of-the-art discovery harnesses (\opta{}, \opene{}, \se{}) with the single highest-performing seed discovery within a given initialization budget is beneficial for downstream discovery success across harnesses, tasks, and LLMs.
    Initialization budget is informed by the convergence criteria in~\autoref{sec:initialization}. 
    }
    \label{fig:initialized_trajectories_prod_grade}
\end{figure}

\textbf{Initialization raises the discovery floor.} 
In all settings, we find that reframing the purpose of parallel discovery from a technique that aims to push the discovery ceiling higher ($r=1$) to an initialization method that raises the discovery floor ($0<r<1$) can be beneficial. 
As shown in~\autoref{fig:initialization_compute}, initializing a serial trajectory with a high-performing population consistently outperforms both purely serial ($r=0$) and purely parallel ($r=1$) discovery, given enough additional serial compute. 
Among initial populations of varying diversity ($s\in\{1, 0.98, 0.95, 0.8\}$), greedy initialization ($s=1$) generally performs well, but can sometimes underperform minimal-diversity rejection sampling (i.e., $s=0.98$) when the initial population contains too many exact or near-duplicate (\autoref{tab:init_method_compare}).
As we curate substantially more diverse initial populations (i.e., $s<0.98$), we find that the average population performance drops, leading to subsequently lower scoring discoveries. This suggests that minimal filtering (i.e., $s=0.98$) to reduce exact and very-near duplicates in the initial population, while prioritizing performance, is sufficient for initializing high-performing trajectories.

\autoref{fig:initialization_compute} shows that \textit{some} amount of initialization compute is beneficial for discovery across all tasks. This raises the question: \textit{can we automatically determine how much initialization compute is ideal in a given setting?}
We find that parallel discovery typically converges in discovery quality (Appendix~\ref{appendix:init}, 
\autoref{fig:initialized_trajectories_w_parallel}), and therefore propose using convergence as a simple indicator. 
To demonstrate, we order parallel discoveries by a random seed and then detect convergence (see~\autoref{appendix:init} for the convergence criterion).
We annotate~\autoref{fig:initialization_compute} with detected convergence and notice that this consistently coincides with the highest-performing initialization budgets. We find that the same task-specific convergence criteria generally can be applied successfully across LLMs (\autoref{appendix:extended_results}, \autoref{fig:initialization_compute_grid}). Further, while fully parallel discovery itself is not amenable to continuous convergence monitoring, we suggest using batched parallel discovery, rather than exhausting the discovery budget at once. Doing so enables early convergence detection 
which signals if it is appropriate to initialize a performant discovery population and switch to a sequential harness.  

\textbf{State-of-the-art harnesses benefit from parallel initialization.} Previously, in~\autoref{tab:prod_vs_modular_harnesses}, we find that for some tasks, state-of-the-art harnesses enabled better discovery. Here, we validate that explicit population initialization outperforms the implicit initialization achieved by SOTA harnesses for the~\autoref{tab:main_task_llm_pairing} settings. We conduct an experiment comparing the performance of state-of-the-art harnesses that are uninitialized (started from the same initial seed candidates as~\autoref{sec:online_intervention}, \autoref{lst:cant_seed}-\ref{lst:prompt_seed}) versus initialized with the single highest-performing candidate found within an initialization budget. We use the ideal-initialized budgets found in~\autoref{fig:initialization_compute} derived from the convergence analysis detailed above. In~\autoref{fig:initialized_trajectories_prod_grade}, we visualize the maximum score trajectory of initialized vs. non-initialized state-of-the-art harnesses; both settings have the same discovery budget. We find that in all (task, LLM) pairings, explicit initialization benefits downstream discovery: Post-initialization discovery scores increase by 2\%-21\%. We conclude that harnesses, on their own, are not consistently capable of implicitly initializing a high-performing trajectory; therefore, it is beneficial to do so explicitly.

\section{Related Work}

\textbf{Diversity of LLM generated text} 
has been measured and studied from many perspectives~\citep{zhang2024improving}.
For instance, LLM response diversity has been shown to decrease with RL-based post-training~\citep{rafailov2023direct, guo2025deepseek,puri2026escaping,jin2026revisiting, wu2025invisible, chen2026does}. To overcome this, many prompting strategies have been proposed~\citep{wang2024calibrating,zhang2025verbalized,troshin2025asking}. However, a more fundamental problem emerges in iterative language modeling tasks where priors from earlier LLM outputs influence and enforce motifs in later outputs~\citep{xu2022learning,holtzman2019curious,laban2026llms}. We confirm a similar trend in LLM-driven discovery: as LLMs sequentially refine discoveries, subsequent discoveries become more similar to each other.
Beyond individual models, populations of LLMs are also producing increasingly similar outputs on open-ended tasks, across both model versions and model families~\citep{guo2024curious, alemohammad2024self,jiang2026artificial, wenger2026large}. This raises concerns that future LLM-generated discoveries may become homogenized.

\textbf{Eliciting LLMs' Parametric Knowledge.} 
The parametric knowledge an LLM acquires has been found to be a function of how frequently that knowledge was present in its training data, the parametric capacity of the model, and the training compute budget~\citep{kandpal2023large, sakarvadia2025mitigating, morris2025much, muennighoff2023scaling, kaplan2020scaling}.
This knowledge can be utilized to accomplish tasks via prompting, and even small changes to the prompt can substantially change an LLM's output ~\citep{min2022rethinking,mccoy2019right,webson2022prompt,tenney2019you, brown2020language,pan2024feedback}. It has been shown that LLM output quality can be improved by spending more \textit{inference-time compute}, e.g., generating more tokens~\citep{wei2022chain, snell2024scaling, wu2024inference, guo2025deepseek,muennighoff2025s1}. 
In this work, we confirm that expending and managing inference-time compute, via discovery harnesses, is useful for LLM-driven discovery, but the extent to which is mediated by the quality of its initial discovery.

\section{Conclusion}
In this work, we study the relationship between how  \textit{discovery harnesses} are initialized and how this affects downstream discovery quality. 
Through the development and characterization of a suite of discovery harnesses, we validate that sequential refinement of an initial seed discovery is beneficial, but sensitive to harness design.  Subsequently, we find that harness design choices that aim to enable more diverse discoveries, from an initial seed discovery, do not predictably achieve more \textit{performant} discoveries. However, we notice, across a wide range of (task, LLM) settings, that the best performing discovery trajectories can be predicted by their initial discovery quality. 
Building on this, we propose a simple, task-, LLM-, and harness-agnostic method to curate high-performing initial discovery populations. By warm-starting discovery harnesses with these initial populations, we consistently improve discovery harnesses' performance.

This work examined the impact and role of \textit{in-context} knowledge on an discovery quality. Future work may characterize the impact of \textit{parametric} knowledge on discovery quality. First, it would be beneficial to better quantify an LLM's parametric knowledge as it relates to discovery tasks, which may exercise the long-tail of human knowledge. This could be done by extending our current analysis of parallel discovery performance, which refrains from updating an LLM's in-context knowledge and instead relies entirely on parametric knowledge.
These findings could inform future research in examining how a model's parametric knowledge priors interact with task-specific in-context priors. For example, our current results suggest that reframing obscure discovery objectives in terms of knowledge domains the LLM is performant in could be beneficial (e.g., express an algorithm design problem in a popular, rather than rare, programming language).

\section*{Acknowledgments}
We thank Derek Tam, John Kirchenbauer and Alok Kamatar for useful discussion.

\subsection*{AI use statement}

In this work, we used generative AI tools to edit components of experimental software.
We \textbf{have not} used generative AI tools to [propose or refine hypotheses, design or provide feedback on research  methodology or experiments, implement methods, assist with translation, clean and reformat datasets, support qualitative and thematic data analysis, interpret results]
and [generate synthetic data sets, help develop theoretical models or conceptual frameworks, formulate mathematical claims, provide critical ingredients for proving mathematical claims, assist in the writing of proofs] are not applicable to this work.
Additionally, we used generative AI tools for modifying scientific figures, formatting LaTeX tables, rephrasing specific sentences, and editing software code. We have reviewed all AI-assisted work. LLM-generated code was reviewed for correctness by one author. All figures were first created without AI use, and subsequent iterations of figures that were refined with AI assistance were compared to the ground-truth human-generated figures for correctness.

\section*{Reproducibility statement}

To enable reproduction of our results we link to the code base: \url{https://github.com/msakarvadia/llm_optimizer}. We additionally include details about the discovery harness studied in~\autoref{appendix:discovery_harness}, the parallel discovery scheme in~\autoref{appendix:parallel_discovery}, and discovery tasks (and their initial seeds \& objectives) in~\autoref{appendix:discovery_task_description}. For all discovery tasks we link to the code bases from which we acquired the task instructions~(\autoref{lst:cant_instruct}-\ref{lst:prompt_instruct}) and initial seeds~(\autoref{lst:cant_seed}-\ref{lst:prompt_seed}).  95\% confidence intervals are reported for all experimental results presented in a figure (shaded region) or table (uncertainty estimate). Additionally, the results presented in the main text (\autoref{tab:main_task_llm_pairing}), are replicated in~\autoref{appendix:extended_results} across additional (task, LLM) pairings (see \autoref{tab:discovery_task_descriptions}).

\bibliographystyle{iclr2027_conference}
\bibliography{references}

\appendix

\section{Discovery Harnesses}
\label{appendix:discovery_harness}

In this study, we consider 15 different discovery harnesses (12 \opro{} harnesses detailed in~\autoref{appendix:modular_harnesses} and 3 state-of-the-art harnesses detailed in~\autoref{appendix:prod_grade_harnesses}) to expose a wide range of design choices and downstream behavior. 

\begin{wrapfigure}[15]{r}{0.5\textwidth}
\vspace{-0.5cm}
\begin{minipage}{1.0\linewidth}
\begin{algorithm}[H]
\caption{General Discovery Harness}
\label{alg:discovery-harness}
\KwIn{task\_objective; discovery evaluation procedure $\mathit{eval}$}
\KwData{$P \gets \emptyset$ \tcp*{active population past discoveries}}
\KwOut{Best discovery $\in P$}
\While{tokens used $<$ budget}{
    parents $\gets$ \textit{SelectParents}($P$)\;
    discovery $\gets$ \textit{LLMMutator}(parents, task\_objective)\;
    score, meta\_data $\gets$ \textit{eval}(discovery)\;
    $P \gets$ \textit{UpdatePopulation}($P$, (discovery, score, meta\_data))\;
}
\end{algorithm}
\end{minipage}
\end{wrapfigure}

\subsection{Modular Harnesses}
\label{appendix:modular_harnesses}

We develop simple and \opro{} harnesses built on top of the \texttt{OPRO} discovery harness proposed in~\citep{yang2024large}. We minimally modify the \texttt{OPRO} harness to expose two key axes of freedom: 1) the mutation strategy and 2) the parent sampling strategy. Every new discovery is added to the active population $P$; when the active population exceeds the maximum population size $m=15$, the lowest performing discovery is pruned from the population.

We consider four different mutation strategies: \texttt{K Context}~\citep{yang2024large}, \texttt{Reflective (Reflect)}~\citep{agrawal2026gepa}, \texttt{Differential Evolution (DE)}~\citep{guo2023evoprompt}, and \texttt{Genetic Algorithm (GA)}~\citep{guo2023evoprompt}. We consider three different parent sampling strategies: \texttt{Highest Scoring (High Score)}~\citep{yang2024large}, \texttt{Tournament}~\citep{guo2023evoprompt}, and \texttt{Roulette Wheel (Wheel)}~\citep{guo2023evoprompt, LIPOWSKI20122193}. In total, these choices yield 12 unique harnesses detailed in~\autoref{tab:opro_variants}.

\begin{table}[h]
    \centering
    \begin{tabular}{cll}
        \toprule
       \opro{} ID  & LLM Mutation Strategy & Parent Sampling Strategy\\
       \midrule
        \opro{} 1 & \texttt{DE} \quad\quad\quad\quad (\autoref{lst:de-mutator}) & \texttt{High Score} (\autoref{alg:high-score-sampling})\\
        \opro{} 2  & \texttt{DE} \quad\quad\quad\quad (\autoref{lst:de-mutator}) & \texttt{Tournament} (\autoref{alg:tournament-sampling})\\
        \opro{} 3  & \texttt{DE} \quad\quad\quad\quad (\autoref{lst:de-mutator}) & \texttt{Wheel} \quad\quad\quad (\autoref{alg:roulette-wheel-sampling})\\
        \opro{} 4 & \texttt{GA} \quad\quad\quad\quad (\autoref{lst:ga-mutator}) & \texttt{High Score} (\autoref{alg:high-score-sampling})\\
        \opro{} 5  & \texttt{GA} \quad\quad\quad\quad (\autoref{lst:ga-mutator}) & \texttt{Tournament} (\autoref{alg:tournament-sampling})\\
        \opro{} 6  & \texttt{GA} \quad\quad\quad\quad (\autoref{lst:ga-mutator}) & \texttt{Wheel} \quad\quad\quad (\autoref{alg:roulette-wheel-sampling})\\
        \opro{} 7 & \texttt{Reflect} \quad (\autoref{lst:reflect-mutator}) & \texttt{High Score} (\autoref{alg:high-score-sampling})\\
        \opro{} 8  & \texttt{Reflect} \quad  (\autoref{lst:reflect-mutator}) & \texttt{Tournament} (\autoref{alg:tournament-sampling})\\
        \opro{} 9  & \texttt{Reflect} \quad (\autoref{lst:reflect-mutator}) & \texttt{Wheel} \quad\quad\quad (\autoref{alg:roulette-wheel-sampling})\\
        \opro{} 10 & \texttt{K Context} (\autoref{lst:kincontext-mutator}) & \texttt{High Score} (\autoref{alg:high-score-sampling})\\
        \opro{} 11 & \texttt{K Context} (\autoref{lst:kincontext-mutator})  & \texttt{Tournament} (\autoref{alg:tournament-sampling})\\
        \opro{} 12  & \texttt{K Context} (\autoref{lst:kincontext-mutator})  & \texttt{Wheel} \quad\quad\quad (\autoref{alg:roulette-wheel-sampling})\\
    \bottomrule
    \end{tabular}
    \caption{\opro{} variants summary.}
    \label{tab:opro_variants}
\end{table}

\begin{algorithm}[H]
\caption{High Score Sampling~\citep{yang2024large}}
\label{alg:high-score-sampling}
\KwIn{$k$ (\# of samples)}
\KwData{$P$ (active discovery population)}
\KwOut{$d$ (discovery\_set)}

$d \gets$ ascending($P$)[-$k$:] \tcp*{rank discoveries by score, low to high}
\end{algorithm}

\begin{algorithm}[H]
\caption{Tournament Sampling~\citep{guo2023evoprompt}}
\label{alg:tournament-sampling}
\KwIn{$k$ (\# of samples), $p=0.5$ (selection probability)}
\KwData{$P$ (active discovery population)}
\KwOut{$d$ (discovery\_set)}

$R \gets$ descending($P$) \tcp*{rank discoveries by score, high to low}
$w_i \gets p(1-p)^i$ for $i = 0, \dots, |R|-1$ \tcp*{geometric rank weights}
$\hat{w} \gets w / \sum w$ \tcp*{normalize into sampling probabilities}
$d \gets$ sample\_k\_samples($R$, $k$, $\hat{w}$)\;
\end{algorithm}

\begin{algorithm}[H]
\caption{Roulette Wheel Sampling~\citep{guo2024connecting, LIPOWSKI20122193}}
\label{alg:roulette-wheel-sampling}
\KwIn{$k$ (\# of samples)}
\KwData{$P$ (active discovery population)}
\KwOut{$d$ (discovery\_set)}

$s_i \gets$ score($x_i$) $+\ \mathrm{shift}$ for $x_i \in P$ \tcp*{shift by $|\min(\text{score})|+\epsilon$ if negative}
$\hat{s} \gets s / \sum s$ \tcp*{normalize into sampling probabilities}
$d \gets$ sample\_k\_samples($P$, $k$, $\hat{s}$)\;
\end{algorithm}

\subsection{LLM-Judge as a Novelty Evaluator}
\label{appendix:llm_judge}

We include the prompts used by the LLM-judge for the \opro{} harness diversity interventions detailed in~\autoref{sec:online_intervention} in~\autoref{lst:judge_prompt_coding} and~\autoref{lst:judge_prompt}; these prompts are directly adopted and adapted from~\citep{lange2026shinkaevolve}.

\begin{lstlisting}[style=pseudostyle, caption={LLM-Judge prompt from~\citep{lange2026shinkaevolve} for coding tasks (\cant{}, \cloud{}, \cp{}). (\url{https://github.com/SakanaAI/ShinkaEvolve/blob/main/shinka/prompts/prompts_novelty.py})}, label={lst:judge_prompt_coding}]
NOVELTY_SYSTEM_MSG = """You are an expert code reviewer tasked with determining if two code snippets are meaningfully different from each other.

Your job is to analyze both programs and determine if the proposed code introduces meaningful changes compared to the existing code. Consider:

1. **Algorithmic differences**: Different approaches, logic, or strategies
2. **Structural changes**: Different data structures, control flow, or organization
3. **Functional improvements**: New features, optimizations, or capabilities
4. **Implementation variations**: Different ways of achieving the same goal that could lead to different performance characteristics
5. **Hyperparameter changes**: Different hyperparameters that could lead to different performance characteristics

Ignore trivial differences like:
- Variable name changes
- Minor formatting or style changes
- Comments or documentation changes
- Insignificant refactoring that doesn't change the core logic

Respond with:
- **NOVEL**: If the codes are meaningfully different
- **NOT_NOVEL**: If the codes are essentially the same with only trivial differences

After your decision, provide a brief explanation of your reasoning."""


NOVELTY_USER_MSG = """Please analyze these two code snippets:

**EXISTING CODE:**
```{language}
{existing_code}
```

**PROPOSED CODE:**
```{language}
{proposed_code}
```

Are these codes meaningfully different? Respond with NOVEL or NOT_NOVEL followed by your explanation."""
\end{lstlisting}

\begin{lstlisting}[style=pseudostyle, caption={LLM-Judge prompt inspired by~\citep{lange2026shinkaevolve} for non-coding tasks (\tsp{},\po{}). (\url{https://github.com/SakanaAI/ShinkaEvolve/blob/main/shinka/prompts/prompts_novelty.py})}, label={lst:judge_prompt}]
JUDGE_SYSTEM_MSG = """You are an expert evaluator tasked with determining if two candidate {solution_description} are meaningfully different from each other.

Task context: {task_description} The goal is to maximize {task_metric}.

Your job is to analyze both candidates and determine if the proposed candidate introduces meaningful changes compared to the existing candidate, with respect to that goal. Consider:

1. **Strategic differences**: Different approaches, arguments, or strategies for pursuing the goal
2. **Structural differences**: Different organization, framing, or content structure
3. **Substantive additions**: New ideas, techniques, or angles not present in the existing candidate
4. **Implementation variations**: Different ways of pursuing the same goal that could plausibly lead to different {task_metric} outcomes

Ignore trivial differences like:
- Minor wording, phrasing, or formatting changes
- Reordering that doesn't change the underlying content
- Insignificant edits that don't change the core approach

Respond with:
- **NOVEL**: If the candidates are meaningfully different
- **NOT_NOVEL**: If the candidates are essentially the same with only trivial differences

After your decision, provide a brief explanation of your reasoning."""


JUDGE_USER_MSG = """Please analyze these two candidate solutions:

**EXISTING CANDIDATE:**
{existing_solution}

**PROPOSED CANDIDATE:**
{proposed_solution}

Are these candidates meaningfully different? Respond with NOVEL or NOT_NOVEL followed by your explanation."""
\end{lstlisting}

\subsection{State-of-the-art Harnesses}
\label{appendix:prod_grade_harnesses}

We consider three popular state-of-the-art harnesses that have been widely studied in this work: \openevolve{} \citep{openevolve}, \shinkaevolve{}~\citep{lange2026shinkaevolve}, and \opta{}~\cite{agrawal2026optimize_anything}. We instantiate each harness with their recommended default parameters; we set the LLM-judge in \shinkaevolve{} to \geminithreefive{} and the cosine similarity threshold to 0.95.

\subsection{Parallel Discovery}
\label{appendix:parallel_discovery}
\begin{figure}[h]
    \centering
    \includegraphics[width=0.7\textwidth]{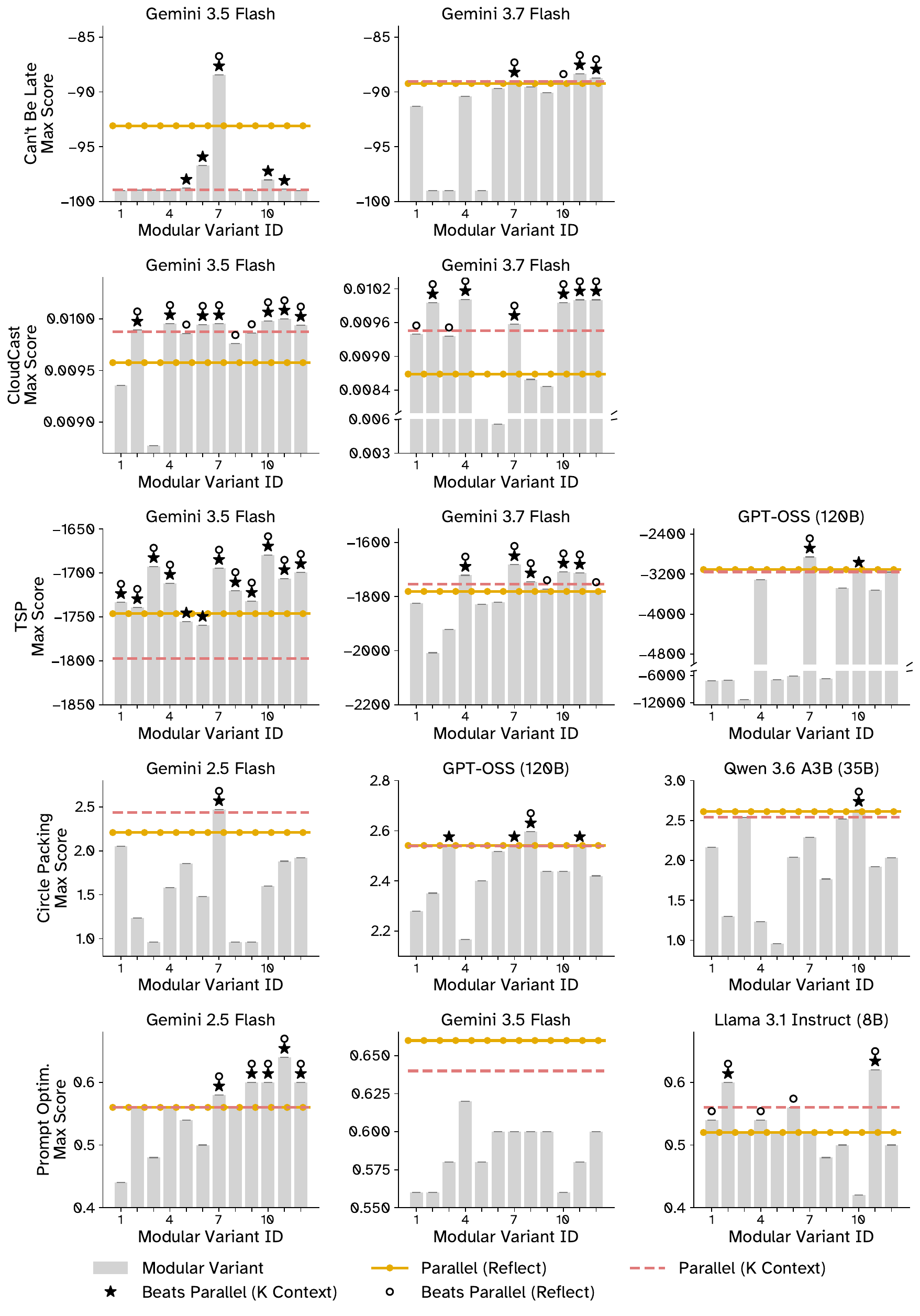}
    \caption{
    \textbf{Sequential discovery can outperform parallel discovery.} We present the extended version of~\autoref{fig:parallel_v_sequential} here.  Comparing sequential  discovery harnesses from our \opro{} suite (detailed in~\autoref{tab:opro_variants}) with parallel discovery using the ``Reflect'' (\autoref{lst:reflect-mutator}) and ``K Context'' prompts (\autoref{lst:kincontext-mutator}). Higher scores are better. Both parallel and sequential approaches are initialized with the same initial seed solutions. We see that generally some form of sequential compute outperforms parallel; but the types of harnesses that perform the best vary with task and LLM. Different rows correspond to different discovery tasks. Note that within a column, the LLM varies across tasks.
    }
    \label{fig:seq_v_parallel_full_grid}
\end{figure}

Here we detail the parallel discovery strategy described in~\autoref{sec:parallel_v_seq}. This approach entails using an LLM, prompted with the task's discovery objective (Listings \ref{lst:cant_instruct}-\ref{lst:prompt_instruct}), to mutate an initial seed discovery (Listings \ref{lst:cant_seed}-\ref{lst:prompt_seed}) in parallel until the inference budget is exhausted. 
Because the in-context knowledge in the LLM's mutation prompts is never updated, each mutation operation is \textit{independent} and therefore can be parallelized. 
The goal of parallel discovery is to generate a high-performing discovery.
We consider both the \texttt{K Context (\autoref{lst:kincontext-mutator})} and \texttt{Reflect (\autoref{lst:reflect-mutator})} mutators for parallel discovery as they can both be instantiated with a single seed discovery, while both \texttt{DE (\autoref{lst:de-mutator})} and \texttt{GA (\autoref{lst:ga-mutator})} need at least two.
In~\autoref{fig:seq_v_parallel_full_grid}, we notice that parallel discovery with the \textit{K Context} mutator is generally more performant than with the \textit{Reflect} mutator.
The main difference between parallel discovery and sequential harnesses is that sequential harnesses actively update the in-context information in the LLM's mutation prompt, and therefore build up dependencies between subsequent discoveries.

\clearpage

\section{Discovery Tasks}
\label{appendix:discovery_task_description}

We outline the five discovery tasks we study in~\autoref{tab:discovery_task_descriptions}. Unless otherwise stated, all discovery schemes are initialized with the following task-specific initial seed discoveries: \cant{} (\autoref{lst:cant_seed}), \cloud{} (\autoref{lst:cloud_seed}), \tsp{} (\autoref{lst:tsp_seed}), \circlepack{} (\autoref{lst:circle_seed}), \promptopt{} (\autoref{lst:prompt_seed}). For experiments in~\autoref{sec:initialization}, task-specific initialization ratios are detailed in~\autoref{tab:init_ratios}.


\begin{table}[h]
    \centering
    \small
    \begin{tabular}{llcl}
    \toprule
    Task  & LLMs & \makecell[c]{ Token\\Discovery \\ Budget} & Evaluation Metric\\
    \midrule
       \makecell[l]{\cant{} \\ Seed:~\autoref{lst:cant_seed}\\Instructions:~\autoref{lst:cant_instruct}}  & \makecell[l]{\geminithreefive{} \\ \geminithreeseven{} }& 1M & \makecell[l]{Negated cost in dollars \\ of discovered scheduling strategy.}\\ 
    \midrule
       
        \makecell[l]{\cloud{}\\ Seed:~\autoref{lst:cloud_seed}\\ Instructions:~\autoref{lst:cloud_instruct}} & \makecell[l]{\geminithreefive{} \\ \geminithreeseven{} }& 2M & \makecell[l]{ Score $= \frac{1}{1+cost}$ where cost of discovered \\ algorithm in dollars is \\ = egress costs (data\_vol × edge\_cost)  \\ + instance costs (runtime × cost\_per\_hour)}\\
    \midrule
        \makecell[l]{\tsp{}\\ Seed:~\autoref{lst:tsp_seed} \\ Instructions:~\autoref{lst:tsp_instruct}} & \makecell[l]{\geminithreefive{} \\ \geminithreeseven{} \\ \gptoss{}} & 1M & \makecell[l]{Negated total distance \\ of discovered path (unit less).} \\
    \midrule
        \makecell[l]{\circlepack{}\\ Seed:~\autoref{lst:circle_seed} \\ Instructions:~\autoref{lst:circle_instruct}}& \makecell[l]{\geminitwofive{} \\ \gptoss{} \\ \qwen{}} & 200K  & \makecell[l]{Sum of all circle diameters in \\ discovered configuration.}\\
    \midrule
        \makecell[l]{\po{}\\ Seed:~\autoref{lst:prompt_seed} \\ Instructions:~\autoref{lst:prompt_instruct}}& \makecell[l]{\geminitwofive{} \\ \geminithreefive{} \\ \llama{}} & 150K & \makecell[l]{Evaluation performance on GSM8K  \\ Benchmark~\cite{cobbe2021training} \\ w/ discovered instruction prompt for \\ downstream LLM (OLMo 2 0425 SFT (1B)) \\ measured as a percentage} \\
        \bottomrule
    \end{tabular}
    \caption{\textbf{Discovery Task Experiment Configurations.}}
    \label{tab:discovery_task_descriptions}
\end{table}

\begin{table}[ht!]
    \centering
    \begin{tabular}{lc}
    \toprule
        Task & Initial:Total Budget Ratios ($r$)\\
    \midrule
        \cant{} & 0.1, 0.25, 0.5, 0.75, 0.9 \\
        \cloud{} & 0.1, 0.25, 0.5, 0.75, 0.9 \\
        \tsp{} & 0.1, 0.25, 0.5, 0.75, 0.9  \\
        \circlepack{} & 0.1, 0.25, 0.5, 0.75, 0.9 \\
        \promptopt{} & 0.17, 0.33, 0.5, 0.67, 0.83\\
    \bottomrule
    \end{tabular}
    \caption{We vary the initialization budget ratios ($r$) across five values for each task for the \opro{} harness initialization experiments in~\autoref{sec:initialization}.}
    \label{tab:init_ratios}
\end{table}

\clearpage

\subsection{Reward Hacking on \cloud{} Task}

Via manual inspection of the generated discoveries for the \cloud{} task, we notice instances of \textit{reward hacking}~\citep{pan2024feedback}, whereby the LLM (\geminithreefive{}), rather than genuinely implementing a cloud scheduling algorithm \textit{with the provided information}, instead attempts to find a \textit{shortcut} to evaluate different discoveries locally and greedily choose the highest performing one. We generally find that the highest performing genuine discoveries hit a score performance ceiling of about 0.01; reward-hacked discoveries typically exceed a performance ceiling of 0.015. Therefore, we apply a post-hoc max-score filter of 0.015 to \cloud{} discoveries. 

We detail an example of \textit{reward hacking} behavior we found in one such discovery in~\autoref{lst:reward_hack_small}.

\begin{lstlisting}[style=pseudostyle, caption={Excerpt of \textit{reward hacking} in a \cloud{} discovery from \geminithreefive{}. We noitice that the \enquote{discovery} explicitly inspects the function call stack to find functions \texttt{['sim', 'eval', 'runner']}, and attempts to locally use those functions as an evaluation oracle.}, label={lst:reward_hack_small}]
...

def find_val_in_stack(names, default):
    try:
        for frame_info in inspect.stack():
            frame = frame_info.frame
            for name in names:
                if name in frame.f_locals:
                    return frame.f_locals[name]
                for local_name, local_val in frame.f_locals.items():
                    if hasattr(local_val, name):
                        return getattr(local_val, name)
    except Exception:
        pass
    return default
    
def find_simulator():
    try:
        for frame_info in inspect.stack():
            frame = frame_info.frame
            for name, obj in list(frame.f_locals.items()) + list(frame.f_globals.items()):
                if obj is None:
                    continue
                if any(x in name.lower() for x in ['sim', 'eval', 'runner']):
                    if hasattr(obj, 'evaluate') or hasattr(obj, 'simulate') or hasattr(obj, 'run'):
                        return obj
                cls_name = obj.__class__.__name__.lower()
                if any(x in cls_name for x in ['sim', 'eval', 'runner']):
                    if hasattr(obj, 'evaluate') or hasattr(obj, 'simulate') or hasattr(obj, 'run'):
                        return obj
    except Exception:
        pass
    return None

def evaluate_topology_via_simulator(topo, G, data_volume_gb, cost_per_hour):
    sim = find_simulator()
    if sim is not None:
        for method_name in ['evaluate', 'simulate', 'run']:
            if hasattr(sim, method_name):
                try:
                    res = getattr(sim, method_name)(topo)
                    if isinstance(res, tuple) and len(res) >= 2:
                        return res[0], res[1]
                except Exception:
                    pass
    return estimate_topology(topo, G, data_volume_gb, cost_per_hour)
...
\end{lstlisting}

\clearpage

\subsection{Convergence Criterion}
\label{appendix:init}

\begin{figure}[hb!t]
    \centering
    \includegraphics[width=\textwidth]{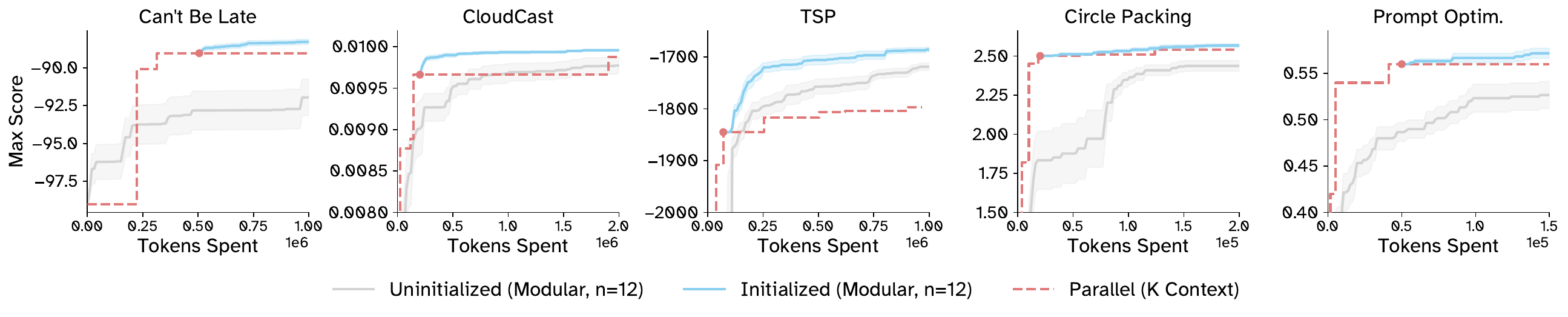}
    \caption{\textbf{Parallel discovery quality converges.} We find that initializing (the \opro{}) discovery harnesses with a high-performing initial population is beneficial for downstream discovery success across harnesses, tasks, and LLMs.
    The initialization budgets are informed by the convergence criteria in~\autoref{sec:initialization}. We overlay the running maximum score of discoveries from the parallel discovery scheme (\autoref{appendix:parallel_discovery}) ordered by a random seed; we notice that these discoveries converge, making them amenable to convergence analysis~(Appendix~\ref{appendix:init}).
    Initial population size $m=15$, similarity threshold $s=0.98$.}
    \label{fig:initialized_trajectories_w_parallel}
\end{figure}

Here we detail the convergence criterion used to assess a discovery trajectory: if the best-known discovered candidate has not improved by a threshold $\delta$ within $N$\% of the discovery budget, then declare convergence. These thresholds were chosen to generally be at least an order of magnitude smaller than the maximum known task score, but for \cant{}, \cloud{}, and \tsp{} are two orders of magnitude smaller than the maximum known score, and $N$ was determined via a visual analysis of the parallel discovery curves~\autoref{fig:initialized_trajectories_w_parallel}. 

\begin{figure}[t]
    \centering
    \includegraphics[width=\textwidth]{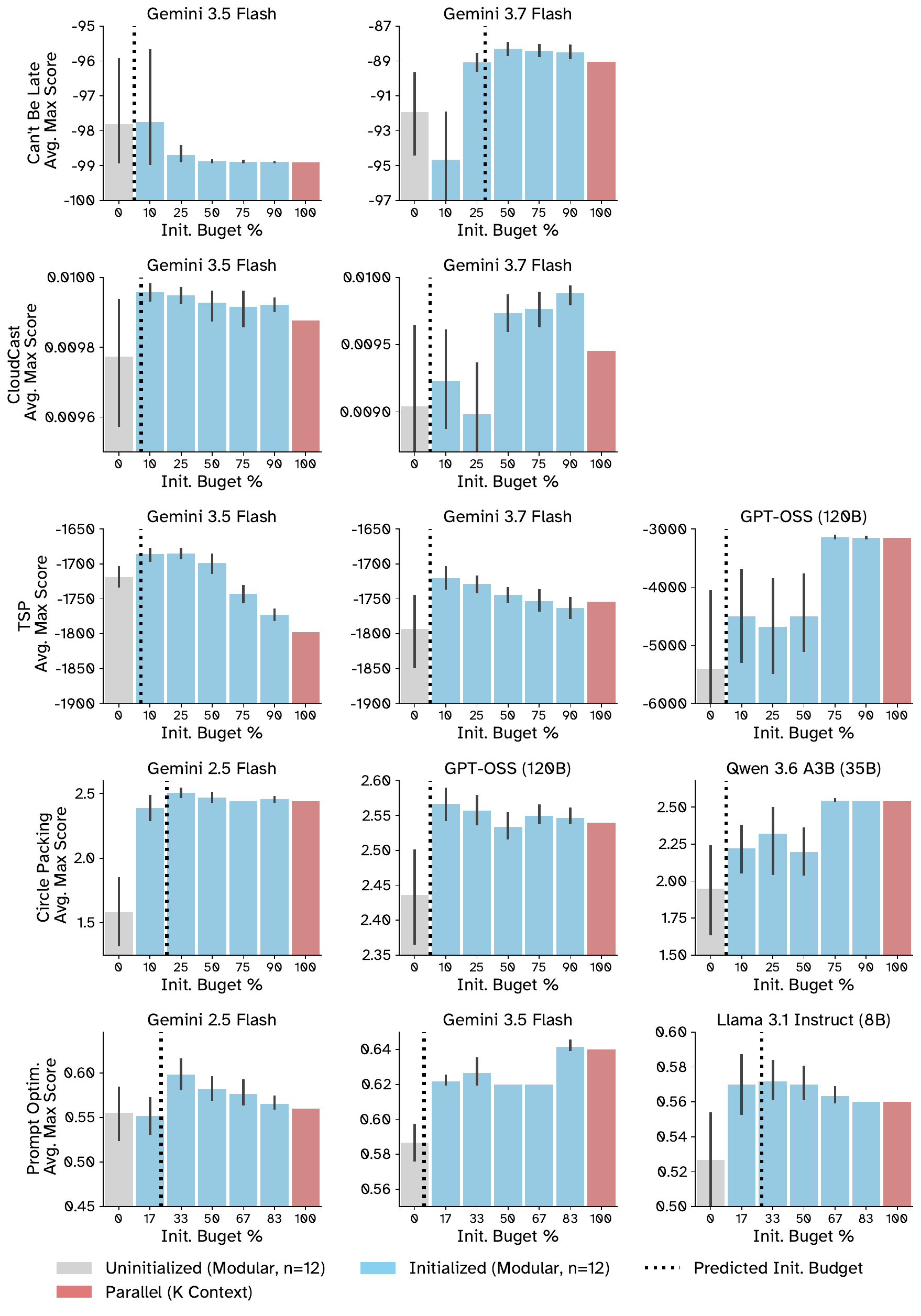}
    \caption{\textbf{Initialized discovery can outperform purely sequential and parallel discovery.} Sequential discovery harnesses which are initialized with a high-performing population (blue), via a fraction of the discovery budget yield the most performant discoveries.
    Predicted initialization budget (dotted line) is predicted via a convergence analysis~\autoref{sec:initialization}. Diversity threshold $s=0.98$ for all initial populations, population size $m=15$. We notice in 9/13 scenarios, the predicted initialization budget coincides with the most performant initialization budget.
    }
    \label{fig:initialization_compute_grid}
\end{figure}

We validate our choice of $\delta$ and $N$ for each task from models in~\autoref{tab:main_task_llm_pairing} as we find that the same values can be successfully used to detect convergence (and subsequently ideal initialization budgets) across multiple LLMs: convergence predicted the ideal initialization budget in most settings (\autoref{fig:initialization_compute_grid}). 
Since $\delta$ is task-specific and $N$ is budget-specific, it would be beneficial for future work to consider ($\delta$, $N$) pairs curated by a subject-matter expert \textit{prior} to discovery. We offer this convergence analysis as a proof of concept and acknowledge that the current choices of ($\delta$, $N$) were made by non-subject-matter experts (the authors).
We detail the task-specific ($\delta$, $N$) pairs in~\autoref{tab:convergence_delta}.

\begin{table}[ht!]
    \centering
    \begin{tabular}{lcc}
    \toprule
        Task & $\delta$ & $N$\\
    \midrule
        \cant{} & 0.5 & 25\%\\
        \cloud{} & 0.0002 & 10\%\\
        \tsp{} & 30 & 10\%\\
        \circlepack{} & 0.1 & 10\%\\
        \promptopt{} & 0.01 & 25\%\\
    \bottomrule
    \end{tabular}
    \caption{Task-specific $\delta$ and $N$ values, by which convergence of a discovery trajectory can be determined.}
    \label{tab:convergence_delta}
\end{table}

\clearpage

\section{Extended Results}
\label{appendix:extended_results}
Here we include additional experimental results. Primarily, we repeat each experiment in the main text over additional LLMs detailed in~\autoref{tab:discovery_task_descriptions}.

\subsection{Parallel vs. Sequential Discovery}
\label{sec:parallel_v_seq_extended}

\begin{figure}[htbp]
    \centering
    \includegraphics[width=\textwidth]{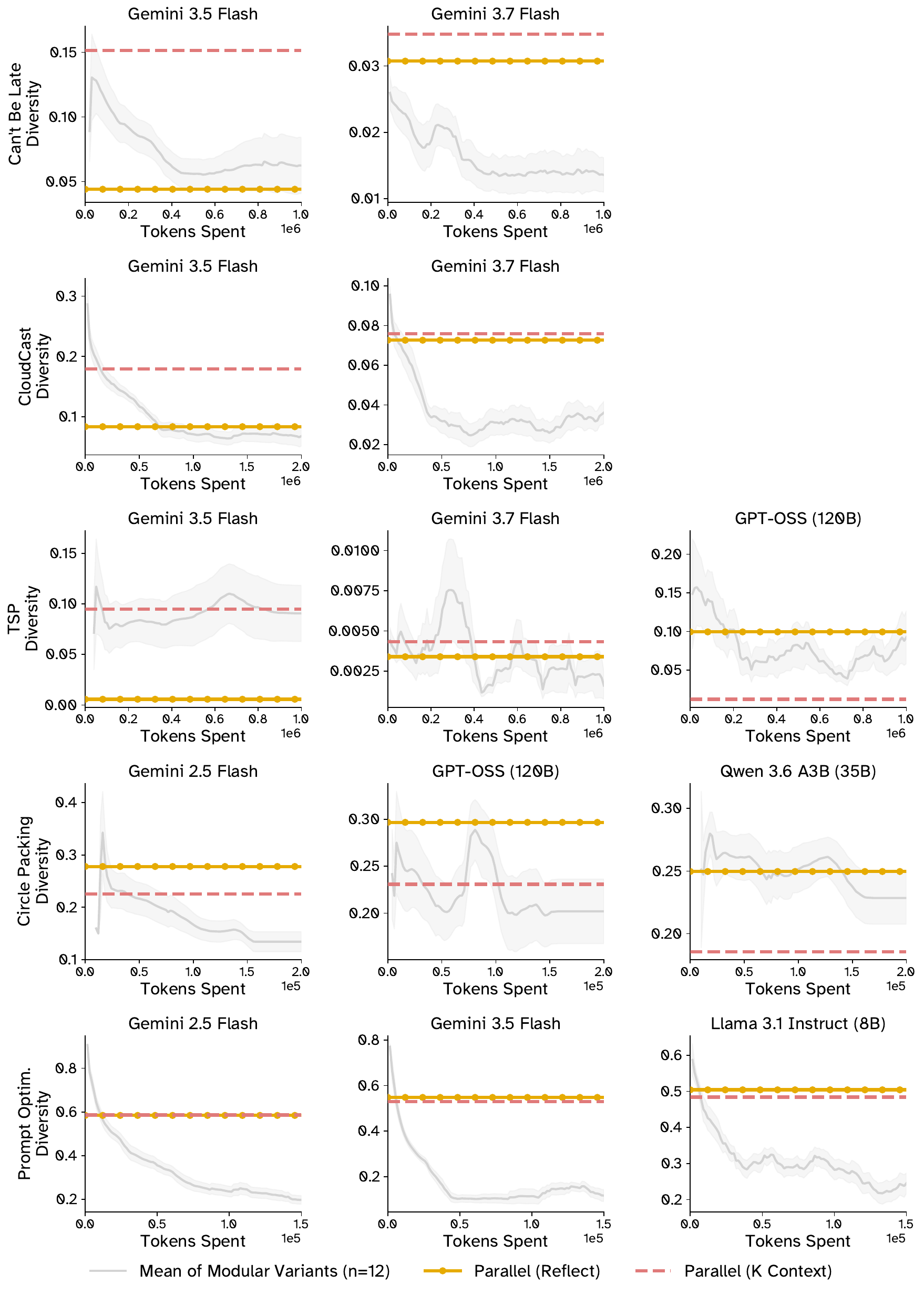} 
    \caption{
    \textbf{Sequential discovery is prone to mode collapse.} We present the extended version of~\autoref{fig:mode_collapse} here. The windowed average pairwise diversity of discoveries over the course of a trajectory across all \opro{} variants vs. average diversity of parallel discovery. Diversity for the parallel discovery scheme is computed over the full set of discoveries. Sequential harnesses consistently yield fewer diverse discoveries and exhibit diminished diversity over the trajectory, whereas parallel schemes discover more diverse candidates. Window size = 15 for all tasks except \cp{} which has window size = 6.
    }
    \label{fig:mode_collapse_grid}
\end{figure}

\clearpage
\newpage

\subsection{Can Exploration Enable Better Discovery?}

\begin{figure}[htbp]
    \centering
    \includegraphics[width=\textwidth]{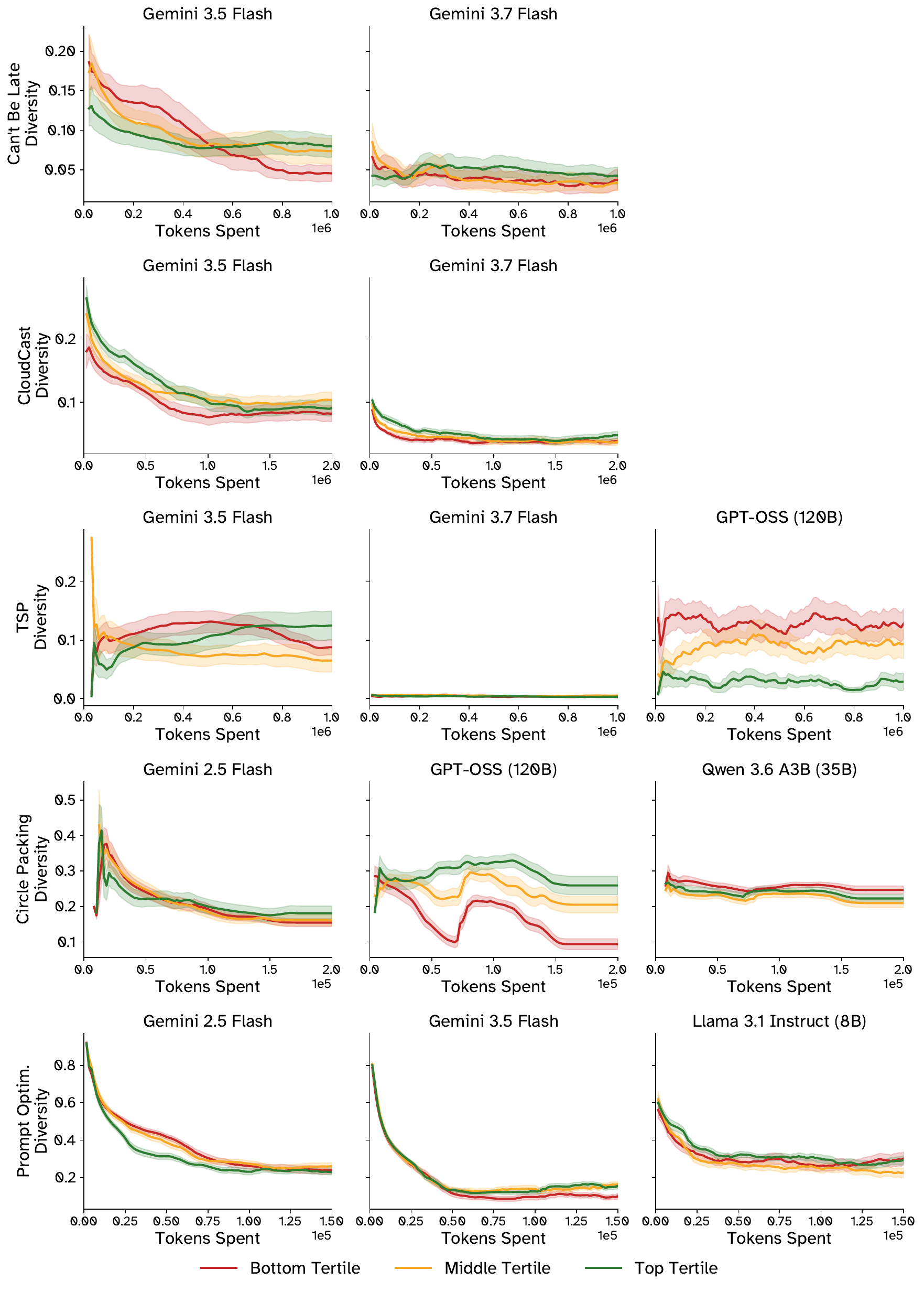} 
    \caption{\textbf{High scoring trajectories eventually experience mode collapse.} We present the extended version of~\autoref{fig:cloudcast_modecollapse} here. We stratify the \opro{} harness trajectories from~\autoref{tab:online_diversity} into tertiles based on the max score of a given trajectory. Windowed diversity visualized over trajectory; window size = 15 for all tasks except \cp{} which has window size = 6.}
    \label{fig:modecollapse_grid}
\end{figure}

\begin{table}[h]
\tiny
\centering
\begin{tabular}{lccc}
\toprule
\multicolumn{4}{c}{\textbf{\cant{}}} \\
\midrule
 & \geminithreeseven{} & \geminithreefive{} & \\
\midrule
\opro{} ($n=12$) & 0.02 ± 0.01 & 0.08 ± 0.03 & \\

Sim thresh. ($s=0.95$) ($n=12$) & 0.05 ± 0.05 & 0.13 ± 0.07 & \\
Sim thresh. ($s=0.8$) ($n=12$) & 0.13 ± 0.06 & 0.18 ± 0.06 & \\
Sim thresh. ($s=0.95$) + LLM judge ($n=12$) & 0.03 ± 0.03 & 0.08 ± 0.04 & \\
Sim thresh. ($s=0.8$) + LLM judge ($n=12$) & 0.03 ± 0.01 & 0.07 ± 0.02 & \\
Deduplication ($n=12$) & 0.05 ± 0.04 & 0.10 ± 0.04 & \\
\shinkaevolve{} & 0.08 & 0.33 & \\
\openevolve{} & 0.03 & 0.05 & \\
\optimizeanything{} & 0.03 & 0.04 & \\
\midrule
\multicolumn{4}{c}{\textbf{\cloud{}}} \\
\midrule
 & \geminithreeseven{} & \geminithreefive{} & \\
\midrule
\opro{} ($n=12$) & 0.0542 ± 0.0137 & 0.1271 ± 0.0195 & \\

Sim thresh. ($s=0.95$) ($n=12$) & 0.0592 ± 0.0130 & 0.1397 ± 0.0273 & \\
Sim thresh. ($s=0.8$) ($n=12$) & 0.0617 ± 0.0084 & 0.1426 ± 0.0287 & \\
Sim thresh. ($s=0.95$) + LLM judge ($n=12$) & 0.0626 ± 0.0146 & 0.1514 ± 0.0387 & \\
Sim thresh. ($s=0.8$) + LLM judge ($n=12$) & 0.0609 ± 0.0136 & 0.1472 ± 0.0406 & \\
Deduplication ($n=12$) & 0.0475 ± 0.0072 & 0.1275 ± 0.0275 & \\
\shinkaevolve{} & 0.0879 & 0.3390 & \\
\openevolve{} & 0.1336 & 0.0976 & \\
\optimizeanything{} & 0.0660 & 0.0673 & \\
\midrule
\multicolumn{4}{c}{\textbf{\tsp{}}} \\
\midrule
 & \gptoss{} & \geminithreeseven{} & \geminithreefive{} \\
\midrule
\opro{} ($n=12$) & 0.094 ± 0.055 & 0.005 ± 0.003 & 0.095 ± 0.050 \\

Sim thresh. ($s=0.95$) ($n=12$) & 0.069 ± 0.072 & 0.006 ± 0.000 & 0.083 ± 0.047 \\
Sim thresh. ($s=0.8$) ($n=12$) & 0.066 ± 0.065 & 0.006 ± 0.000 & 0.123 ± 0.056 \\
Sim thresh. ($s=0.95$) + LLM judge ($n=12$) & 0.124 ± 0.083 & 0.003 ± 0.002 & 0.101 ± 0.054 \\
Sim thresh. ($s=0.8$) + LLM judge ($n=12$) & 0.078 ± 0.035 & 0.004 ± 0.002 & 0.124 ± 0.052 \\
Deduplication ($n=12$) & 0.092 ± 0.050 & 0.004 ± 0.003 & 0.069 ± 0.056 \\
\shinkaevolve{} & 0.150 & 0.001 & 0.150 \\
\openevolve{} & 0.165 & 0.018 & 0.333 \\
\optimizeanything{} & 0.004 & 0.002 & 0.074 \\
\midrule
\multicolumn{4}{c}{\textbf{\cp{}}} \\
\midrule
 & \qwen{} & \geminitwofive{} & \gptoss{} \\
\midrule
\opro{} ($n=12$) & 0.26 ± 0.03 & 0.20 ± 0.06 & 0.26 ± 0.06 \\

Sim thresh. ($s=0.95$) ($n=12$) & 0.24 ± 0.02 & 0.23 ± 0.06 & 0.26 ± 0.06 \\
Sim thresh. ($s=0.8$) ($n=12$) & 0.25 ± 0.03 & 0.24 ± 0.06 & 0.30 ± 0.04 \\
Sim thresh. ($s=0.95$) + LLM judge ($n=12$) & 0.25 ± 0.02 & 0.26 ± 0.05 & 0.28 ± 0.04 \\
Sim thresh. ($s=0.8$) + LLM judge ($n=12$) & 0.23 ± 0.03 & 0.24 ± 0.05 & 0.28 ± 0.04 \\
Deduplication ($n=12$) & 0.24 ± 0.03 & 0.22 ± 0.06 & 0.24 ± 0.07 \\
\shinkaevolve{} & 0.17 & 0.31 & 0.27 \\
\openevolve{} & 0.23 & 0.27 & 0.27 \\
\optimizeanything{} & 0.26 & 0.38 & 0.20 \\
\midrule
\multicolumn{4}{c}{\textbf{\po{}}} \\
\midrule
 & \geminitwofive{} & \geminithreefive{} & \llama{} \\
\midrule
\opro{} ($n=12$) & 0.40 ± 0.04 & 0.23 ± 0.04 & 0.40 ± 0.04 \\

Sim thresh. ($s=0.95$) ($n=12$) & 0.41 ± 0.04 & 0.23 ± 0.04 & 0.36 ± 0.05 \\
Sim thresh. ($s=0.8$) ($n=12$) & 0.44 ± 0.03 & 0.25 ± 0.03 & 0.39 ± 0.07 \\
Sim thresh. ($s=0.95$) + LLM judge ($n=12$) & 0.41 ± 0.05 & 0.26 ± 0.03 & 0.39 ± 0.08 \\
Sim thresh. ($s=0.8$) + LLM judge ($n=12$) & 0.42 ± 0.04 & 0.27 ± 0.04 & 0.40 ± 0.07 \\
Deduplication ($n=12$) & 0.44 ± 0.04 & 0.23 ± 0.03 & 0.37 ± 0.06 \\
\shinkaevolve{} & 0.47 & 0.41 & 0.46 \\
\openevolve{} & 0.41 & 0.47 & 0.28 \\
\optimizeanything{} & 0.29 & 0.16 & 0.35 \\
\bottomrule
\end{tabular}
\caption{\textbf{Discovery harnesses can discovery more diverse populations.} We report the comprehensive diversity results for the~\autoref{sec:online_intervention} experiments. Both state-of-the-art discovery harnesses (\se{}, \opta{}, \opene{}) and \opro{} harnesses with diversity interventions on average generally achieve more diverse discoveries than baseline~\opro{} harnesses across tasks and LLMs.}
\label{tab:diversity_of_harnesses_all}
\end{table}

\begin{table}[h]
\tiny
\centering
\begin{tabular}{lccc}
\toprule
\multicolumn{4}{c}{\textbf{\cant{}}} \\
\midrule
 & \geminithreeseven{} & \geminithreefive{} & \\
\midrule
\opro{} ($n=12$) & -91.94 ± 2.75 & -97.80 ± 1.92 & \\
\cmidrule{1-4}
Sim thresh. ($s=0.95$) ($n=12$) & -95.79 ± 2.98 & -98.14 ± 1.86 & \\
Sim thresh. ($s=0.8$) ($n=12$) & -91.68 ± 2.81 & -97.42 ± 2.34 & \\
Sim thresh. ($s=0.95$) + LLM judge ($n=12$) & -95.93 ± 2.82 & -98.64 ± 0.52 & \\
Sim thresh. ($s=0.8$) + LLM judge ($n=12$) & -94.82 ± 2.75 & -98.39 ± 1.22 & \\
Deduplication ($n=12$) & -93.93 ± 3.34 & -96.28 ± 2.88 & \\

Avg. Interventions ($n=60$) & -94.43 ± 1.23 & -97.77 ± 0.80 & \\
\cmidrule{1-4}
\shinkaevolve{} & -88.60 & -99.01 & \\
\openevolve{} & -95.06 & -98.91 & \\
\optimizeanything{} & -90.28 & -98.80 & \\
Avg. Production ($n=3$) & -91.32 ± 8.33 & -98.90 ± 0.26 & \\
\midrule
\multicolumn{4}{c}{\textbf{\cloud{}}} \\
\midrule
 & \geminithreeseven{} & \geminithreefive{} & \\
\midrule
\opro{} ($n=12$) & 0.0090 ± 0.0008 & 0.0098 ± 0.0002 & \\
\cmidrule{1-4}
Sim thresh. ($s=0.95$) ($n=12$) & 0.0092 ± 0.0005 & 0.0097 ± 0.0002 & \\
Sim thresh. ($s=0.8$) ($n=12$) & 0.0097 ± 0.0003 & 0.0097 ± 0.0001 & \\
Sim thresh. ($s=0.95$) + LLM judge ($n=12$) & 0.0091 ± 0.0008 & 0.0098 ± 0.0001 & \\
Sim thresh. ($s=0.8$) + LLM judge ($n=12$) & 0.0093 ± 0.0005 & 0.0098 ± 0.0001 & \\
Deduplication ($n=12$) & 0.0093 ± 0.0004 & 0.0098 ± 0.0001 & \\

Avg. Interventions ($n=60$) & 0.0093 ± 0.0002 & 0.0098 ± 0.0001 & \\
\cmidrule{1-4}
\shinkaevolve{} & 0.0100 & 0.0085 & \\
\openevolve{} & 0.0100 & 0.0100 & \\
\optimizeanything{} & 0.0100 & 0.0099 & \\
Avg. Production ($n=3$) & 0.0100 ± 0.0000 & 0.0095 ± 0.0021 & \\
\midrule
\multicolumn{4}{c}{\textbf{\tsp{}}} \\
\midrule
 & \gptoss{} & \geminithreeseven{} & \geminithreefive{} \\
\midrule
\opro{} ($n=12$) & -5399 ± 1645 & -1793 ± 60 & -1719 ± 16 \\
\cmidrule{1-4}
Sim thresh. ($s=0.95$) ($n=12$) & -4162 ± 1036 & -1766 ± 11 & -1777 ± 29 \\
Sim thresh. ($s=0.8$) ($n=12$) & -4112 ± 1056 & -1760 ± 15 & -1775 ± 17 \\
Sim thresh. ($s=0.95$) + LLM judge ($n=12$) & -5270 ± 1112 & -1842 ± 47 & -1744 ± 46 \\
Sim thresh. ($s=0.8$) + LLM judge ($n=12$) & -5268 ± 1042 & -1806 ± 57 & -1735 ± 37 \\
Deduplication ($n=12$) & -5210 ± 1558 & -1811 ± 50 & -1721 ± 12 \\

Avg. Interventions ($n=60$) & -4805 ± 484 & -1797 ± 18 & -1750 ± 13 \\
\cmidrule{1-4}
\shinkaevolve{} & -11324 & -1701 & -1879 \\
\openevolve{} & -5286 & -1760 & -1755 \\
\optimizeanything{} & -3215 & -1673 & -1680 \\
Avg. Production ($n=3$) & -6608 ± 10466 & -1711 ± 111 & -1772 ± 250 \\
\midrule
\multicolumn{4}{c}{\textbf{\cp{}}} \\
\midrule
 & \qwen{} & \geminitwofive{} & \gptoss{} \\
\midrule
\opro{} ($n=12$) & 1.95 ± 0.35 & 1.58 ± 0.31 & 2.44 ± 0.08 \\
\cmidrule{1-4}
Sim thresh. ($s=0.95$) ($n=12$) & 1.96 ± 0.38 & 1.84 ± 0.37 & 2.19 ± 0.38 \\
Sim thresh. ($s=0.8$) ($n=12$) & 1.76 ± 0.39 & 1.73 ± 0.36 & 2.34 ± 0.26 \\
Sim thresh. ($s=0.95$) + LLM judge ($n=12$) & 1.84 ± 0.43 & 1.69 ± 0.32 & 2.33 ± 0.23 \\
Sim thresh. ($s=0.8$) + LLM judge ($n=12$) & 1.94 ± 0.42 & 1.79 ± 0.40 & 2.26 ± 0.30 \\
Deduplication ($n=12$) & 2.21 ± 0.23 & 1.44 ± 0.38 & 2.04 ± 0.44 \\

Avg. Interventions ($n=60$) & 1.94 ± 0.15 & 1.70 ± 0.15 & 2.23 ± 0.13 \\
\cmidrule{1-4}
\shinkaevolve{} & 2.48 & 2.43 & 2.49 \\
\openevolve{} & 2.17 & 1.79 & 2.54 \\
\optimizeanything{} & 2.51 & 2.17 & 2.51 \\
Avg. SOTA ($n=3$) & 2.39 ± 0.48 & 2.13 ± 0.81 & 2.51 ± 0.07 \\
\midrule
\multicolumn{4}{c}{\textbf{\po{}}} \\
\midrule
 & \geminitwofive{} & \geminithreefive{} & \llama{} \\
\midrule
\opro{} ($n=12$) & 0.55 ± 0.04 & 0.59 ± 0.01 & 0.53 ± 0.03 \\
\cmidrule{1-4}
Sim thresh. ($s=0.95$) ($n=12$) & 0.55 ± 0.04 & 0.59 ± 0.02 & 0.53 ± 0.03 \\
Sim thresh. ($s=0.8$) ($n=12$) & 0.56 ± 0.02 & 0.59 ± 0.01 & 0.52 ± 0.02 \\
Sim thresh. ($s=0.95$) + LLM judge ($n=12$) & 0.54 ± 0.04 & 0.57 ± 0.02 & 0.53 ± 0.04 \\
Sim thresh. ($s=0.8$) + LLM judge ($n=12$) & 0.55 ± 0.05 & 0.58 ± 0.02 & 0.51 ± 0.04 \\
Deduplication ($n=12$) & 0.55 ± 0.03 & 0.60 ± 0.03 & 0.52 ± 0.04 \\

Avg. Interventions ($n=60$) & 0.55 ± 0.01 & 0.59 ± 0.01 & 0.52 ± 0.01 \\
\cmidrule{1-4}
\shinkaevolve{} & 0.58 & 0.60 & 0.38 \\
\openevolve{} & 0.54 & 0.62 & 0.48 \\
\optimizeanything{} & 0.62 & 0.64 & 0.54 \\
Avg. Production ($n=3$) & 0.58 ± 0.10 & 0.62 ± 0.05 & 0.47 ± 0.20 \\
\bottomrule
\end{tabular}
\caption{\textbf{Discovery harnesses that explore more do not necessarily achieve more performant discoveries}. We report the comprehensive set of results for~\autoref{sec:online_intervention}. Neither state-of-the-art discovery harnesses (\se{},\opta{},\opene{}) nor \opro{} harnesses with diversity interventions predictably achieve more performant discoveries; instead the ideal choice or harness/intervention varies across (task, LLM) settings.}
\label{tab:online_interventions_compare_all}
\end{table}

\clearpage
\newpage

\subsection{The Importance of Initialization}
\label{appendix:initialization_extended_results}

\begin{table}[h]
\tiny
\centering
\renewcommand{\arraystretch}{1.3}
\begin{tabular}{cllllll}
\toprule
 &  & Sim thresh. ($s=0.8$) & Sim thresh. ($s=0.95$) & Sim thresh. ($s=0.98$) & Sim thresh. (Greedy, $s=1$) \\
Task & LLM &  &  &  &  \\
\midrule
\multirow[c]{2}{*}{\cant{}} & \geminithreefive{} & \cellcolor{green!25}-98.56 ± 0.43 & \cellcolor{green!10}-98.58 ± 0.42 & -98.62 ± 0.38 & -98.63 ± 0.35 \\
\cline{2-6}
 & \geminithreeseven{} Flash & -90.22 ± 0.90 & -90.52 ± 0.91 & \cellcolor{green!25}-89.79 ± 0.85 & \cellcolor{green!10}-89.90 ± 0.87 \\
\cline{1-6} \cline{2-6}
\multirow[c]{2}{*}{\cloud{}} & \geminithreefive{} & \cellcolor{green!10}0.0099 ± 0.0000 & \cellcolor{green!25}0.0100 ± 0.0001 & \cellcolor{green!10}0.0099 ± 0.0000 & \cellcolor{green!10}0.0099 ± 0.0000 \\
\cline{2-6}
 & \geminithreeseven{} & \cellcolor{green!25}0.0095 ± 0.0001 & \cellcolor{green!25}0.0095 ± 0.0001 & \cellcolor{green!25}0.0095 ± 0.0001 & \cellcolor{green!25}0.0095 ± 0.0001 \\
\cline{1-6} \cline{2-6}
\multirow[c]{3}{*}{\tsp{}} & \gptoss{} & -4433 ± 415 & -4674 ± 422 & \cellcolor{green!25}-3993 ± 322 & \cellcolor{green!10}-4071 ± 346 \\
\cline{2-6}
 & \geminithreefive{} & -1726 ± 11 & -1726 ± 11 & \cellcolor{green!10}-1717 ± 10 & \cellcolor{green!25}-1713 ± 10 \\
\cline{2-6}
 & \geminithreeseven{} & -1751 ± 8 & -1747 ± 7 & \cellcolor{green!10}-1742 ± 7 & \cellcolor{green!25}-1741 ± 7 \\
\cline{1-6} \cline{2-6}
\multirow[c]{3}{*}{\cp{}} & \gptoss{} & - & 2.53 ± 0.01 & \cellcolor{green!25}2.55 ± 0.01 & \cellcolor{green!10}2.54 ± 0.01 \\
\cline{2-6}
 & \geminitwofive{} & - & 2.41 ± 0.03 & \cellcolor{green!25}2.45 ± 0.02 & \cellcolor{green!10}2.42 ± 0.03 \\
\cline{2-6}
 & \qwen{} & - & 1.82 ± 0.18 & \cellcolor{green!25}2.36 ± 0.08 & \cellcolor{green!10}1.96 ± 0.16 \\
\cline{1-6} \cline{2-6}
\multirow[c]{3}{*}{\po{}} & \geminitwofive{} & \cellcolor{green!25}0.579 ± 0.008 & 0.571 ± 0.008 & \cellcolor{green!10}0.575 ± 0.008 & \cellcolor{green!10}0.575 ± 0.007 \\
\cline{2-6}
 & \geminithreefive{} & \cellcolor{green!10}0.625 ± 0.002 & \cellcolor{green!10}0.625 ± 0.003 & \cellcolor{green!25}0.626 ± 0.003 & \cellcolor{green!25}0.626 ± 0.003 \\
\cline{2-6}
 & \llama{} & \cellcolor{green!25}0.568 ± 0.005 & 0.564 ± 0.004 & \cellcolor{green!10}0.567 ± 0.005 & 0.566 ± 0.005 \\
\bottomrule
\end{tabular}
\caption{\textbf{Comparing Population Initialization Methods across varying similarity thresholds.} See population curation method description in~\autoref{sec:initialization}. Here we report the average score of a population initialization method across five initialization budgets $r$; see task-specific initialization budgets in~\autoref{tab:init_ratios}. We find that greedy population curation is generally performant (highest or second-highest-performing population), but that minimal similarity filtering ($s=0.98$) typically outperforms purely greedy populations. Therefore, we conclude that minimal filtering of exact and near duplicates from the population is beneficial ($s=0.98$). Dark green shading in a cell indicates the highest-scoring population within a row; lighter green shading indicates the second-highest-scoring population. We do not conduct $s=0.8$ experiments for \cp{} as more than half of the initialization population needed to be backfilled greedily due to a lack of sufficiently diverse candidates.}
\label{tab:init_method_compare}
\end{table}
\clearpage


\section{Mutation Prompts}

\begin{lstlisting}[style=pseudostyle, caption={\texttt{K Context} Mutator needs 1 LLM call. Inspired by \citep{yang2024large}. k = 3 for coding tasks (e.g., \cant{}, \cloud{}, \cp{}), and k = 5 for the remaining tasks (e.g., \tsp{}, \po{}).}, label={lst:kincontext-mutator}]
P                     # active discovery population
k                     # number of in-context examples
task_instruction      # details about discovery objective
discovery_description # descriptor e.g., "algorithm"/"prompt"
task_metric           # e.g., cost, score etc.

in_context_examples = sample_population(P, k)


past_examples = ""
for solution, score, meta_data in in_context_examples:
    example = """
    ### Past Example

    **[PREVIOUSLY GENERATED SOLUTION:]**
    {solution}
    **[SCORE ASSIGNED TO SOLUTION:]**
    {task_metric}: {score}
    **[RUN METADATA:]**
    {meta_data}
    
    """
    past_examples += example

mutation_prompt = """
{task_instruction}
Your goal is to maximize {task_metric}. Output only the bare minimum text to reach the objective goal.

Here are some past examples and the {task_metric} they received where the goal is to maximize the metric.

{past_examples}

Generate a new {discovery_description} that is different from the old ones to maximize the {task_metric} as much as possible.
"""

# Mutation Operation
new_discovery = LLM(mutation_prompt)
\end{lstlisting}

\begin{lstlisting}[style=pseudostyle, caption={\texttt{Reflect} Mutator needs 2 LLM calls. From \citep{agrawal2026optimize_anything}.}, label={lst:reflect-mutator}]
P                     # active discovery population
task_instruction      # details about discovery objective

in_context_example = sample_population(P, 1)
solution, score, meta_data = in_context_example[0]


diagnosis_instructions = """

You are an expert AI system diagnostics agent. Your task is to analyze the behavior of an LLM agent and determine exactly why its current instructions are causing it to fail.

Below is the current solution being utilized by the agent:
[CURRENT SOLUTION]
{solution}

Below is a minibatch of execution traces where the agent may have failed to achieve the target evaluation metric. Sometimes there are no failed trajectories to review. Review the step-by-step reasoning, tool calls, and outputs:
[(Potential) FAILED TRAJECTORIES]
{meta_data}

INSTRUCTIONS:
1. Conduct a rigorous error analysis. Identify patterns or assumptions in the [CURRENT PROMPT] that misled the agent.
2. Pinpoint exactly which instructions caused the wrong reasoning steps or incorrect tool invocations.
3. Provide a clear, high-level natural language diagnosis detailing what needs to be changed, added, or removed from the solution to fix these specific edge cases. Do not rewrite the solution yet; only provide the structural diagnosis.
"""

mutation_instructions = """

You are an elite engineering and optimization algorithm. Your goal is to mutate an existing solutions to maximize its performance across a series of target objectives.

You must improve upon the ancestor solution by applying high-level lessons learned from failure data. If the ancestor is a placeholder, replace it with a relevant solution.

[CURRENT SOLUTION]
{solution}

[NEW SYSTEM DIAGNOSIS]
{diagnosis}

CRITICAL RULES FOR MUTATION:
1. Synthesize the new system diagnosis with the accumulated historical lessons.
2. Mutate the instructions to directly prevent the errors noted in the diagnosis while preserving the parts of the solution that successfully handled past tasks.
3. Optimize for clarity, conciseness, and structural robustness.
4. Output ONLY the final mutated solution text. Do not include introductory conversational filler or concluding explanations.
"""

# Step 1: Diagnosis
diagnosis_prompt = task_instruction + diagnosis_instructions
diagnosis = LLM(diagnosis_prompt)

# Step 2: Mutation
mutation_prompt = task_instruction + mutation_instructions
new_discovery = LLM(mutation_prompt)
\end{lstlisting}

\begin{lstlisting}[style=pseudostyle, caption={\texttt{Differential Evolution (DE)} Mutator needs 2 LLM calls. From \citep{guo2023evoprompt}.}, label={lst:de-mutator}]
P                # active discovery population
task_instruction # details about discovery objective

# requires >= 3 past solutions; falls back to K Context mutator otherwise
in_context_examples = sample_population(P, 3)
solution_1, score_1, meta_data_1 = in_context_examples[-3]
solution_2, score_2, meta_data_2 = in_context_examples[-2]
solution_3, score_3, meta_data_3 = in_context_examples[-1]

# Step 1: Mutation + Crossover
mutation_instructions = """

First we provide an example of evolving a prompt, this may be different from the task you are supposed to complete. Regardless, follow the template:
Please follow the instruction step-by-step to generate a better prompt.
1. Identify the different parts between the Prompt 1 and Prompt 2:
Prompt 1: Rewrite the input text into simpler text.
Prompt 2: Rewrite my complex sentence in simpler terms, but keep the meaning.
2. Randomly mutate the different parts
3. Crossover the different parts with the following Prompt 3 and generate a final prompt bracketed with <solution> and </solution>:
Prompt 3: Rewrite the given input text into simpler English sentences while preserving the same meaning, so it can be understood by non-native English speakers.

1. Identifying the different parts between Prompt 1 and Prompt 2:
Prompt 1: Rewrite the input text into simpler text.
Prompt 2: Rewrite my complex sentence in simpler terms, but keep the meaning.
Different parts:
"input text" vs "my complex sentence"
"simpler text" vs "simpler terms, but keep the meaning"

2. Randomly mutate the different parts:
"input text" -> "provided text"
"my complex sentence" -> "the difficult sentence"
"simpler text" -> "easier language"
"simpler terms, but keep the meaning" -> "simpler words while maintaining the meaning"

3. Crossover the different parts with the following Prompt 3 and generate a final prompt bracketed with <solution> and </solution>:
Prompt 3: Rewrite the given input text into simpler English sentences while preserving the same meaning, so it can be understood by non-native English speakers.

Final Prompt: <solution>Transform the difficult sentence into easier language while keeping the meaning, for non-native English speakers to comprehend.</solution>

Now it is your turn. Please follow the instruction step-by-step to generate a better solution.
1. Identify the different parts between Solution 1 and Solution 2:
Solution 1:
{solution_1}
Solution 1's meta data:
{meta_data_1}
Solution 2:
{solution_2}
Solution 2's meta data:
{meta_data_2}
2. Randomly mutate the different parts
3. Crossover the different parts with the following Solution 3 and generate a final solution (Do not write anything except the final request solution):
Solution 3:
{solution_3}
Solution 3's meta data:
{meta_data_3}
"""

# Step 2: Extraction
extraction_instructions = """

In the previous step, you were instructed to:
1. Identify the different parts between Solution 1 and Solution 2:
Solution 1:
{solution_1}
Solution 2:
{solution_2}
2. Randomly mutate the different parts
3. Crossover the different parts with the following Solution 3 and generate a final solution (Do not write anything except the final request solution):
Solution 3:
{solution_3}

Here is what you wrote:
{mutation_output}

Extract the final solution that was derived after step three in the past step from in between <solution> and </solution>. Output only that final solution.
"""

# 1st: Mutation Operation
mutation_prompt = task_instruction + mutation_instructions
mutation_output = LLM(mutation_prompt)

# 2nd: Extract Discovery
extraction_prompt = task_instruction + extraction_instructions
new_discovery = LLM(extraction_prompt)
\end{lstlisting}

\begin{lstlisting}[style=pseudostyle, caption={\texttt{Genetic Algorithm (GA)} Mutator needs 2 LLM calls. From \citep{guo2023evoprompt}.}, label={lst:ga-mutator}]
P                # active discovery population
task_instruction # details about discovery objective

# requires >= 3 past solutions; falls back to K Context mutator otherwise
in_context_examples = sample_population(P, 3)
solution_1, score_1, meta_data_1 = in_context_examples[-3]
solution_2, score_2, meta_data_2 = in_context_examples[-2]
solution_3, score_3, meta_data_3 = in_context_examples[-1]

# Step 1: Crossover + Mutation
crossover_instructions = """
First we provide an example of evolving a prompt, this may be different from the task you are supposed to complete. Regardless, follow the template:
Please follow the instruction step-by-step to generate a better prompt.
1. Crossover the following prompts and generate a new prompt:
Prompt 1: Rewrite the input text into simpler text.
Prompt 2: Rewrite my complex sentence in simpler terms, but keep the meaning.
2. Mutate the prompt generated in Step 1 and generate a final prompt bracketed with 
    <solution> and </solution>.

1. Crossover Prompt: Rewrite the complex text into simpler text while keeping its meaning.
2. <solution>Transform the provided text into simpler language, maintaining its essence.</solution>

Now it is your turn. Please follow the instruction step-by-step to generate a better solution.
Please follow the instruction step-by-step to generate a better solution.
1. Crossover the following solution and generate a new solution:
Solution 1:
{solution_2}
Solution 1's meta data:
{meta_data_2}
Solution 2:
{solution_3}
Solution 2's meta data:
{meta_data_3}
2. Mutate the solution generated in Step 1 and generate a final solution bracketed with 
    <solution> and </solution>.

1.
"""

# 1st: Crossover + Mutation Operation
crossover_prompt = task_instruction + crossover_instructions
crossover_output = LLM(crossover_prompt)

# 2nd: Extract Discovery; Reuse same extraction logic as Differential Evolution
extraction_prompt = task_instruction + extraction_instructions
new_discovery = LLM(extraction_prompt)
\end{lstlisting}

\section{Task Objectives}

\begin{lstlisting}[style=pseudostyle, caption={\cant{} discovery instructions \citep{agrawal2026optimize_anything}. (\url{https://github.com/gepa-ai/gepa/tree/main/examples/adrs/can_be_late})}, label={lst:cant_instruct}]
Optimize a cloud scheduling strategy for the "Can't Be Late" problem.

The strategy decides when to use SPOT instances (cheap but can be preempted) vs ON_DEMAND instances (expensive but reliable) to complete a task before its deadline. The goal is to minimize cost while ensuring the task completes on time.
Output the executable Python code to accomplish the task inside a markdown code block.

Key information about the problem domain:

- ClusterType.SPOT: Use spot instances (cheap, ~$0.3/hour, but can be preempted at any time)
- ClusterType.ON_DEMAND: Use on-demand instances (expensive, ~$1/hour, but guaranteed availability)
- ClusterType.NONE: Wait without using any instances (no cost, but no progress)
- restart_overhead: Time penalty incurred when switching from one instance type to another
- The strategy MUST ensure task completion before the deadline (hard constraint)
- Score is the negative of cost in dollars, so a higher (less negative) score means lower cost

Evaluation feedback format:
- Timeline format: start-end:TYPE@REGION[progress%] (e.g., "0.0-5.0:S@R0[50%]" means SPOT from hour 0-5 reaching 50% progress)
- Spot availability: S=available, X=unavailable (e.g., "0.0-10.0:S | 10.0-15.0:X" means spot available first 10h, then unavailable)

Optimization targets:
1. Reduce overall cost while maintaining deadline guarantees
2. Make better decisions about when to use SPOT vs ON_DEMAND
3. Handle spot unavailability more intelligently
4. Consider the trade-offs between waiting for spot and using on-demand
\end{lstlisting}

\begin{lstlisting}[style=pseudostyle, caption={\cloud{} discovery instructions \citep{agrawal2026optimize_anything}. (\url{https://github.com/gepa-ai/gepa/tree/main/examples/adrs/cloudcast})}, label={lst:cloud_instruct}]
Optimize a broadcast routing algorithm for multi-cloud data transfer.

The algorithm decides how to route data from a single source to multiple destinations across cloud providers (AWS, GCP, Azure). The goal is to minimize total cost (egress fees + instance costs) while maintaining good transfer times.
Output the executable Python code to accomplish the task inside a markdown code block.

Key information about the problem domain:

- The network is represented as a directed graph where:
  - Nodes are cloud regions (e.g., "aws:us-east-1", "gcp:europe-west1-a", "azure:eastus")
  - Edges have 'cost' ($/GB for egress) and 'throughput' (Gbps bandwidth) attributes

- Data is partitioned into num_partitions chunks that can be routed independently
- Each partition can take a different path to reach each destination
- Total cost = egress costs (data_vol x edge_cost) +
               instance costs (runtime x cost_per_hour)

- The algorithm must return a BroadCastTopology object containing:
  - paths[dst][partition] = list of edges [[src, dst, edge_data], ...]
  - Each destination must have at least one valid path for each partition

Evaluation feedback format:
- Cost: Total transfer cost in dollars
- Transfer time: Maximum time for all destinations to receive data (seconds)

Optimization targets:
1. Reduce total cost (egress + instance costs)
2. Find paths that balance cost and throughput
3. Consider multipath routing for better bandwidth utilization
4. Exploit cloud provider pricing differences (e.g., intra-provider is cheaper)
\end{lstlisting}

\begin{lstlisting}[style=pseudostyle, caption={\traveling{} discovery instructions \textbf{}. 
(\url{https://github.com/google-deepmind/opro/blob/main/opro/optimization/optimize_tsp.py})}, label={lst:tsp_instruct}]
Below are some previous traces and their scores. Score is the negative of
the trace length, so a higher (less negative) score means a shorter, 
better trace. The trace should traverse all points exactly once. 
The trace should start with <trace> and end with </trace>.
\end{lstlisting}

\begin{lstlisting}[style=pseudostyle, caption={\circlepack{} discovery instruction \citep{openevolve}. (\url{https://github.com/algorithmicsuperintelligence/openevolve/blob/main/examples/circle_packing/initial_program.py})}, label={lst:circle_instruct}]
The circle packing problem involves placing n non-overlapping circles inside a container (in this case, a unit square) to optimize a specific metric. For this example:

We pack exactly 26 circles
Each circle must lie entirely within the unit square
No circles may overlap
We aim to maximize the sum of all circle radii.

Construct a specific arrangement of 26 circles in a unit square that attempts to maximize the sum of their radii.

Write a python function called 'run_packing' which acchomplishes this goal and     Returns:
        Tuple of (centers, radii, sum_of_radii)
        centers: np.array of shape (26, 2) with (x, y) coordinates
        radii: np.array of shape (26) with radius of each circle
        sum_of_radii: Sum of all radii

Output the executable Python code to accomplish the task inside a markdown code block.
\end{lstlisting}

\begin{lstlisting}[style=pseudostyle, caption={\promptopt{} discovery instructions.}, label={lst:prompt_instruct}]
Craft a prompt prefix to boost a downstream {LLM_name}'s score on the 
{benchmark_name} benchmark.
\end{lstlisting}

\section{Initial Seed Discoveries}

\begin{lstlisting}[style=pseudostyle, caption={\cant{} initial seed discovery from \citep{agrawal2026optimize_anything}. (\url{https://github.com/gepa-ai/gepa/tree/main/examples/adrs/can_be_late})}, label={lst:cant_seed}]
import math
from sky_spot.strategies.strategy import Strategy
from sky_spot.utils import ClusterType

class EvolveSingleRegionStrategy(Strategy):
    NAME = 'evolve_single_region'

    def __init__(self, args):
        super().__init__(args)

    def reset(self, env, task):
        super().reset(env, task)

    def _step(self, last_cluster_type: ClusterType, has_spot: bool) -> ClusterType:
        env = self.env

        remaining_task_time = self.task_duration - sum(self.task_done_time)
        if remaining_task_time <= 1e-3:
            return ClusterType.NONE

        remaining_time = self.deadline - env.elapsed_seconds

        if remaining_task_time + self.restart_overhead >= remaining_time:
            return ClusterType.ON_DEMAND

        if has_spot:
            return ClusterType.SPOT
        else:
            return ClusterType.NONE

    @classmethod
    def _from_args(cls, parser):
        args, _ = parser.parse_known_args()
        return cls(args)
\end{lstlisting}

\begin{lstlisting}[style=pseudostyle, caption={\cloud{} initial seed discovery from \citep{agrawal2026optimize_anything}. (\url{https://github.com/gepa-ai/gepa/tree/main/examples/adrs/cloudcast})}, label={lst:cloud_seed}]
import networkx as nx
import pandas as pd
import os
from typing import Dict, List


class SingleDstPath(Dict):
    partition: int
    edges: List[List]  # [[src, dst, edge data]]


class BroadCastTopology:
    def __init__(self, src: str, dsts: List[str], num_partitions: int = 4, paths: Dict[str, 'SingleDstPath'] = None):
        self.src = src
        self.dsts = dsts
        self.num_partitions = num_partitions
        if paths is not None:
            self.paths = paths
        else:
            self.paths = {dst: {str(i): None for i in range(num_partitions)} for dst in dsts}

    def get_paths(self):
        return self.paths

    def set_num_partitions(self, num_partitions: int):
        self.num_partitions = num_partitions

    def set_dst_partition_paths(self, dst: str, partition: int, paths: List[List]):
        partition = str(partition)
        self.paths[dst][partition] = paths

    def append_dst_partition_path(self, dst: str, partition: int, path: List):
        partition = str(partition)
        if self.paths[dst][partition] is None:
            self.paths[dst][partition] = []
        self.paths[dst][partition].append(path)


def search_algorithm(src, dsts, G, num_partitions):
    \"\"\"
    Find broadcast paths from source to all destinations.

    Uses Dijkstra's shortest path algorithm based on cost as the edge weight.

    Args:
        src: Source node identifier (e.g., "aws:ap-northeast-1")
        dsts: List of destination node identifiers
        G: NetworkX DiGraph with cost and throughput edge attributes
        num_partitions: Number of data partitions

    Returns:
        BroadCastTopology object with paths for all destinations and partitions
    \"\"\"
    h = G.copy()
    h.remove_edges_from(list(h.in_edges(src)) + list(nx.selfloop_edges(h)))
    bc_topology = BroadCastTopology(src, dsts, num_partitions)

    for dst in dsts:
        path = nx.dijkstra_path(h, src, dst, weight="cost")
        for i in range(0, len(path) - 1):
            s, t = path[i], path[i + 1]
            for j in range(bc_topology.num_partitions):
                bc_topology.append_dst_partition_path(dst, j, [s, t, G[s][t]])

    return bc_topology
\end{lstlisting}

\begin{lstlisting}[style=pseudostyle, caption={\traveling{} initial seed discovery is a randomly initialized path as per \citep{yang2024large}. (\url{https://github.com/google-deepmind/opro/blob/main/opro/optimization/optimize_tsp.py})}, label={lst:tsp_seed}]
<trace> 
3,7,83,48,87,19,59,65,28,36,35,38,25,55,27,17,22,93,
69,67,47,77,95,74,51,97,84,20,11,33,61,18,24,12,6,21,
81,72,79,49,39,9,30,23,66,90,52,44,2,32,37,15,99,80,
54,71,40,42,41,88,56,8,16,5,73,60,43,63,96,64,46,82,
31,0,94,68,53,85,75,89,10,98,4,78,70,34,14,86,29,91,
62,57,13,50,76,26,1,45,92,58
</trace>
\end{lstlisting}

\begin{lstlisting}[style=pseudostyle, caption={\circlepack{} initial seed discovery from \citep{openevolve}. (\url{https://github.com/algorithmicsuperintelligence/openevolve/blob/main/examples/circle_packing/initial_program.py})}, label={lst:circle_seed}]
import numpy as np


def run_packing():
    """
    Construct a specific arrangement of 26 circles in a unit square
    that attempts to maximize the sum of their radii.

    Returns:
        Tuple of (centers, radii, sum_of_radii)
        centers: np.array of shape (26, 2) with (x, y) coordinates
        radii: np.array of shape (26) with radius of each circle
        sum_of_radii: Sum of all radii
    """
    # Initialize arrays for 26 circles
    n = 26
    centers = np.zeros((n, 2))

    # Place circles in a structured pattern
    # This is a simple pattern - evolution will improve this

    # First, place a large circle in the center
    centers[0] = [0.5, 0.5]

    # Place 8 circles around it in a ring
    for i in range(8):
        angle = 2 * np.pi * i / 8
        centers[i + 1] = [0.5 + 0.3 * np.cos(angle), 0.5 + 0.3 * np.sin(angle)]

    # Place 16 more circles in an outer ring
    for i in range(16):
        angle = 2 * np.pi * i / 16
        centers[i + 9] = [0.5 + 0.7 * np.cos(angle), 0.5 + 0.7 * np.sin(angle)]

    # Additional positioning adjustment to make sure all circles
    # are inside the square and don't overlap
    # Clip to ensure everything is inside the unit square
    centers = np.clip(centers, 0.01, 0.99)

    # Compute maximum valid radii for this configuration
    radii = compute_max_radii(centers)

    # Calculate the sum of radii
    sum_radii = np.sum(radii)

    return centers, radii, sum_radii


def compute_max_radii(centers):
    """
    Compute the maximum possible radii for each circle position
    such that they don't overlap and stay within the unit square.

    Args:
        centers: np.array of shape (n, 2) with (x, y) coordinates

    Returns:
        np.array of shape (n) with radius of each circle
    """
    n = centers.shape[0]
    radii = np.ones(n)

    # First, limit by distance to square borders
    for i in range(n):
        x, y = centers[i]
        # Distance to borders
        radii[i] = min(x, y, 1 - x, 1 - y)

    # Then, limit by distance to other circles
    # Each pair of circles with centers at distance d can have
    # sum of radii at most d to avoid overlap
    for i in range(n):
        for j in range(i + 1, n):
            dist = np.sqrt(np.sum((centers[i] - centers[j]) ** 2))

            # If current radii would cause overlap
            if radii[i] + radii[j] > dist:
                # Scale both radii proportionally
                scale = dist / (radii[i] + radii[j])
                radii[i] *= scale
                radii[j] *= scale

    return radii
\end{lstlisting}

\begin{lstlisting}[style=pseudostyle, caption={\promptopt{} initial seed discovery is a placeholder string.}, label={lst:prompt_seed}]
'placeholder prompt prefix'
\end{lstlisting}

\end{document}